\documentclass[letterpaper]{article} 
\usepackage{aaai25}  
\usepackage{times}  
\usepackage{helvet}  
\usepackage{courier}  
\usepackage[hyphens]{url}  
\usepackage{graphicx} 
\usepackage{natbib}  
\usepackage{caption} 
\usepackage{algorithm}
\usepackage{amsmath} 
\usepackage{booktabs}
\usepackage{multirow}
\usepackage{gensymb}
\usepackage{algpseudocode} 
\usepackage{pgfplots}
\usepackage{makecell}
\usepackage{subcaption}
\usepackage{amsfonts}
\usepackage{float}
\usepackage{appendix}
\newcommand{\noi}[1]{\noindent\textbf{#1}}

\usepackage{newfloat}
\usepackage{listings}
\DeclareCaptionStyle{ruled}{labelfont=normalfont,labelsep=colon,strut=off} 
\floatstyle{ruled}
\newfloat{listing}{tb}{lst}{}
\floatname{listing}{Listing}
\title{GoHD: Gaze-oriented and Highly Disentangled Portrait Animation with Rhythmic Poses and Realistic Expressions}
\author{
    Ziqi Zhou\textsuperscript{\rm 1,2},
    Weize Quan\textsuperscript{\rm 1,2},
    Hailin Shi\textsuperscript{\rm 3},
    Wei Li\textsuperscript{\rm 4},
    Lili Wang\textsuperscript{\rm 5},
    Dong-Ming Yan\textsuperscript{\rm 1,2}\footnote{Corresponding author: yandongming@gmail.com.}
}
\affiliations{
    \textsuperscript{\rm 1}MAIS, Institute of Automation, Chinese Academy of Sciences
    \textsuperscript{\rm 2}University of Chinese Academy of Sciences\\
    \textsuperscript{\rm 3}NIO
    \textsuperscript{\rm 4}Banma
    \textsuperscript{\rm 5}State Key Laboratory of Virtual Reality Technology and Systems, Beihang University
}

\usepackage{bibentry}

\begin{document} 
\maketitle
\begin{abstract}
Audio-driven talking head generation necessitates seamless integration of audio and visual data amidst the challenges posed by diverse input portraits and intricate correlations between audio and facial motions. In response, we propose a robust framework GoHD designed to produce highly realistic, expressive, and controllable portrait videos from any reference identity with any motion. GoHD innovates with three key modules: Firstly, an animation module utilizing latent navigation is introduced to improve the generalization ability across unseen input styles. This module achieves high disentanglement of motion and identity, and it also incorporates gaze orientation to rectify unnatural eye movements that were previously overlooked. Secondly, a conformer-structured conditional diffusion model is designed to guarantee head poses that are aware of prosody. Thirdly, to estimate lip-synchronized and realistic expressions from the input audio within limited training data, a two-stage training strategy is devised to decouple frequent and frame-wise lip motion distillation from the generation of other more temporally dependent but less audio-related motions, e.g., blinks and frowns. Extensive experiments validate GoHD's advanced generalization capabilities, demonstrating its effectiveness in generating realistic talking face results on arbitrary subjects. Our implementation is available at https://github.com/Jia1018/GoHD.
\end{abstract} 
\section{Introduction}
\label{sec:intro}
Audio-driven portrait animation, widely applied in social media and mixed reality contexts like avatar creation and teleconferencing, has made notable progress fueled by artificial intelligence \cite{chen2019hierarchical, zhou2020makelttalk, wang2021audio2head, wav2lip, Zhang_2023_CVPR, yu2023thpad, tian2024emo, xu2024vasa1, drobyshev2024emoportraits}. However, various problems persist in existing animation methods. Specifically, some struggle with maintaining natural mouth shapes when animating exaggerated expressions \cite{zhou2020makelttalk, wav2lip, Zhang_2023_CVPR}, while others encounter severe warping distortions and identity alternations for unseen data \cite{wang2021audio2head, eamm}. In addition to the difficulties in generating audio-synchronized lip motions, there are challenges in accurately estimating other spontaneous motions like head poses and eye motions, often resulting in poor performance \cite{zhou2020makelttalk, Zhang_2023_CVPR} or reliance on another reference video \cite{zhou2021pose, eamm, styletalk, ma2023dreamtalk}, which is not available in most scenarios. Consequently, crafting a robust portrait animation framework that is effective for all types of input portraits and can independently generate satisfactory talking motions remains an unresolved issue.
\begin{figure}
    \centering
    \includegraphics[width=\linewidth]{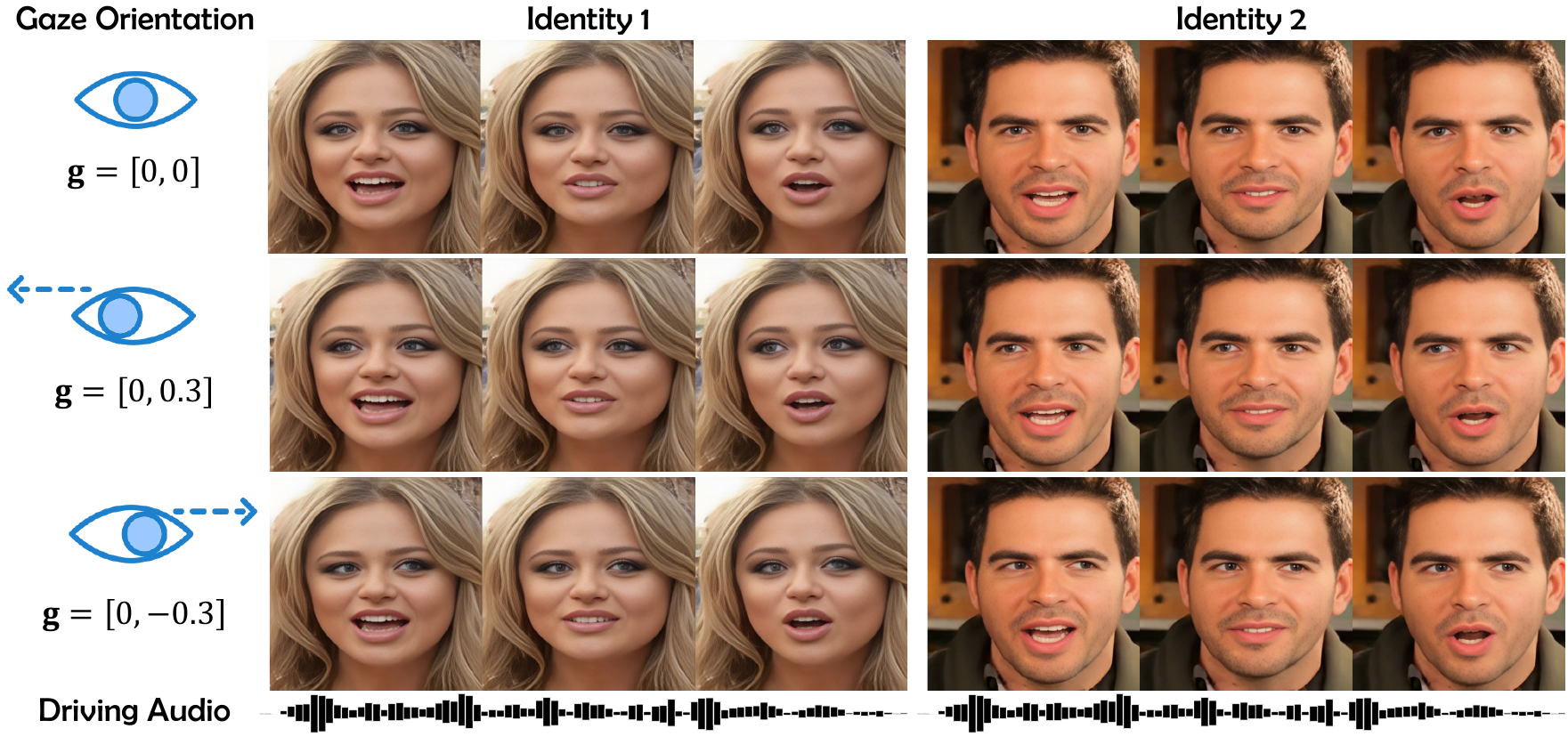}
    \caption{Illustration of gaze orientation experiments. The results of two identities driven by the same audio clip and different gaze directions are presented. The true pitch and yaw angles are multiplied by $\pi$.}
    \label{fig:gaze}
    \vspace{-0.3cm}
\end{figure}

Therefore, to devise a novel talking face system that can generalize well to any initial identities with various facial motions, several challenges remain to be addressed: 1) The input portraits may vary significantly in terms of appearance, expression, and other factors, requiring the system to learn a robust representation of facial features and movements that can be applied to new, unseen subjects. Animating techniques of current methods \cite{wang2021audio2head, eamm, zhou2020makelttalk} fail to fully disentangle identity and motion, resulting in poor generalization and distortions, especially when applied to out-of-distribution images with rich expressions. 2) Existing implicit or explicit motion representation methods \cite{deng2019accurate, zhou2021pose, yu2023thpad} for facial animation encounter limitations in gaze orientation, inevitably creating unnatural looking directions in the generated videos. 3) The intricate mappings between audio and low-frequency motions (e.g. head poses) require models to incorporate both prosody awareness and result diversity. Prior works that use probabilistic methods \cite{Zhang_2023_CVPR} or sequence-to-sequence models \cite{wang2021audio2head} often emphasize one aspect—either diversity or prosody—while ignoring a balanced consideration of both features. 4) Learning precise lip-audio alignment requires enormous training sample pairs to achieve cross-modal adaption in the feature spaces, which is often inaccessible to regular researchers. Additionally, the interplay between mouth movements and other spontaneous facial actions (less correlated with audio), such as blinks and frowns, can introduce complexity to the overall expression generation process. 

To overcome these difficulties, we propose \textbf{GoHD}, a \textbf{G}aze-\textbf{o}riented and \textbf{H}ighly \textbf{D}isentangled portrait animation method with audio-driven rhythmic head poses and realistic facial expressions. Specifically, GoHD is composed of three main modules: a generalized latent navigable face animator with gaze orientation, a prosody-aware denoising network for pose generation, and an expression estimator trained in a two-stage manner. Firstly, to accomplish fully disentangled motion transformation for arbitrary input identity, we integrate the animation module with latent navigation techniques \cite{wang2022latent}, skillfully decoupling a latent motion space from the underlying identity. More precisely, we split it into a source branch and a driving branch. In the source branch, a latent identity code is generated for each input reference image, representing the appearance feature without any head poses or expressions. Meanwhile, the driving branch processes target motions as inputs and predict a motion vector based on a learned motion codebook. Gaze directions are incorporated as conditions in this branch to provide overall motion control and rectify potential unnatural eye movements. The animated result is then obtained by decoding a combined representation of the predicted motion vector and the identity code.

Additionally, to realize audio-driven and controllable portrait animation, we design two independent generators for the driving motions in the face animator. An audio-conditioned diffusion model with a conformer-based denoising network is used to map audio cues to head poses, capturing prosody patterns with dilated convolutions and self-attention modules for natural, sequential results. The great probabilistic sampling characteristic \cite{ho2020denoising, alexanderson2023listen, kong2021diffwave, shen2023difftalk} of diffusion models further enhance the diversity of generated outputs. Regarding expressions, we focus on audio-related eye and lip motions, where lip movements require precise frame-wise synchronization, and eye motions like blinks and frowns depend more on temporal dynamics. To bridge this gap, we extract handcrafted eye motion features from pre-defined expression coefficients and introduce an audio-to-expression prediction approach trained by a two-stage strategy. The first stage focuses on distilling precise frame-wise lip motions from an expert pre-trained on sufficient audio-visual pairs \cite{wav2lip}, while the second stage uses an LSTM (Long Short-Term Memory) structured model to generate temporally dependent eye motions. With our well-designed two-stage training scheme, realistic and audio-synchronized expression generation is achieved with effective disentanglement of lip and eye motions.

In summary, this paper contributes in the following ways: 1) We propose a gaze-oriented and robust face animation module using latent navigation that effectively disentangles motion from identity. 2) We present a conformer-based conditional diffusion model for generating rhythmic and realistic poses. 3) A two-stage training strategy for expression prediction is devised to bridge the frequency gap between lip and eye motions. 4) Extensive experiments demonstrate that our method can generate advanced talking face results on arbitrary subjects with the proposed motion generation and animation modules.

\section{Related Work}
\label{sec:related_work}
\noi{Audio-driven Talking Face Animation.}
The goal of this task is to generate a video where the input face image animates in synchronization with the provided audio. 
Early approaches \cite{chung2017you, Vougioukas2019EndtoEndSR, ijcai2019p129} adopt end-to-end networks for direct frame-wise generation from input face image and audio. To enhance audio-visual control, Chen et al. \shortcite{chen2019hierarchical} uses explicit facial landmarks, while Zhou et al. \shortcite{zhou2018talking} employs disentangled latent representations. PC-AVS \cite{zhou2021pose} addresses spontaneous motions like head poses with a decoupled latent pose space. StyleTalk \cite{styletalk} introduces a style-controllable decoder, and Yu et al. \cite{yu2023thpad} decompose the latent space into lip and non-lip spaces. Some other works \cite{wang2021audio2head, wang2021one} independently predict head motions but can lead to face distortion and identity alternation. MakeItTalk \cite{zhou2020makelttalk} estimates speaker-specific motions with facial landmarks, limiting expression conveyance. MODA \cite{liu2023MODA} enhances motion decoupling with denser landmarks. Later works \cite{Ren_2021_ICCV, zhang2021flow, Zhang_2023_CVPR} explore 3DMMs, but appear desynchronized lip motions \cite{Ren_2021_ICCV, zhang2021flow} and unrealistic poses \cite{Zhang_2023_CVPR}. More recently, the world's famous AI labs released several outstanding works \cite{he2023gaia, xu2024vasa1, drobyshev2024emoportraits, tian2024emo} in this area, yet their requirements for huge training datasets are not practical to regular researchers. Our work introduces a novel framework capable of generating more realistic overall facial motions while addressing the practical challenge of limited training data availability.

\noi{Video-driven Talking Face Motion Imitation.}
In this category, the objective is to create a new video where the source face image adeptly mimics the expressions and head movements of the input driving video. 
Intermediaries are crucial for precise motion transformation. FOMM \cite{Siarohin_2019_NeurIPS} uses learned key points and their affine transformations as structural references, while methods \cite{wang2021facevid2vid, siarohin2021motion, hong2022depth, zhao2022thin} enhance it with 3D \cite{wang2021facevid2vid} or depth information \cite{hong2022depth}, and modified motion estimation \cite{siarohin2021motion, zhao2022thin}. In contrast, LIA \cite{wang2022latent} introduces a motion warping framework, navigating in latent space to avoid errors from explicit representations. Pang et al. \cite{Pang_2023_CVPR} extend this approach with bidirectional cyclic training for disentangled pose and expression editing. StyleHEAT \cite{yin2022styleheat} uses a pre-trained StyleGAN \cite{karras2020analyzing} for high-resolution motion driving and editing, but may lead to identity discrepancies and artifacts. 
Our method adopts a latent navigable approach \cite{wang2022latent} for simple and effective motion transformation in animating talking faces with predicted coefficients.

\section{Method}
\begin{figure}
    \centering
    \includegraphics[width=\linewidth]{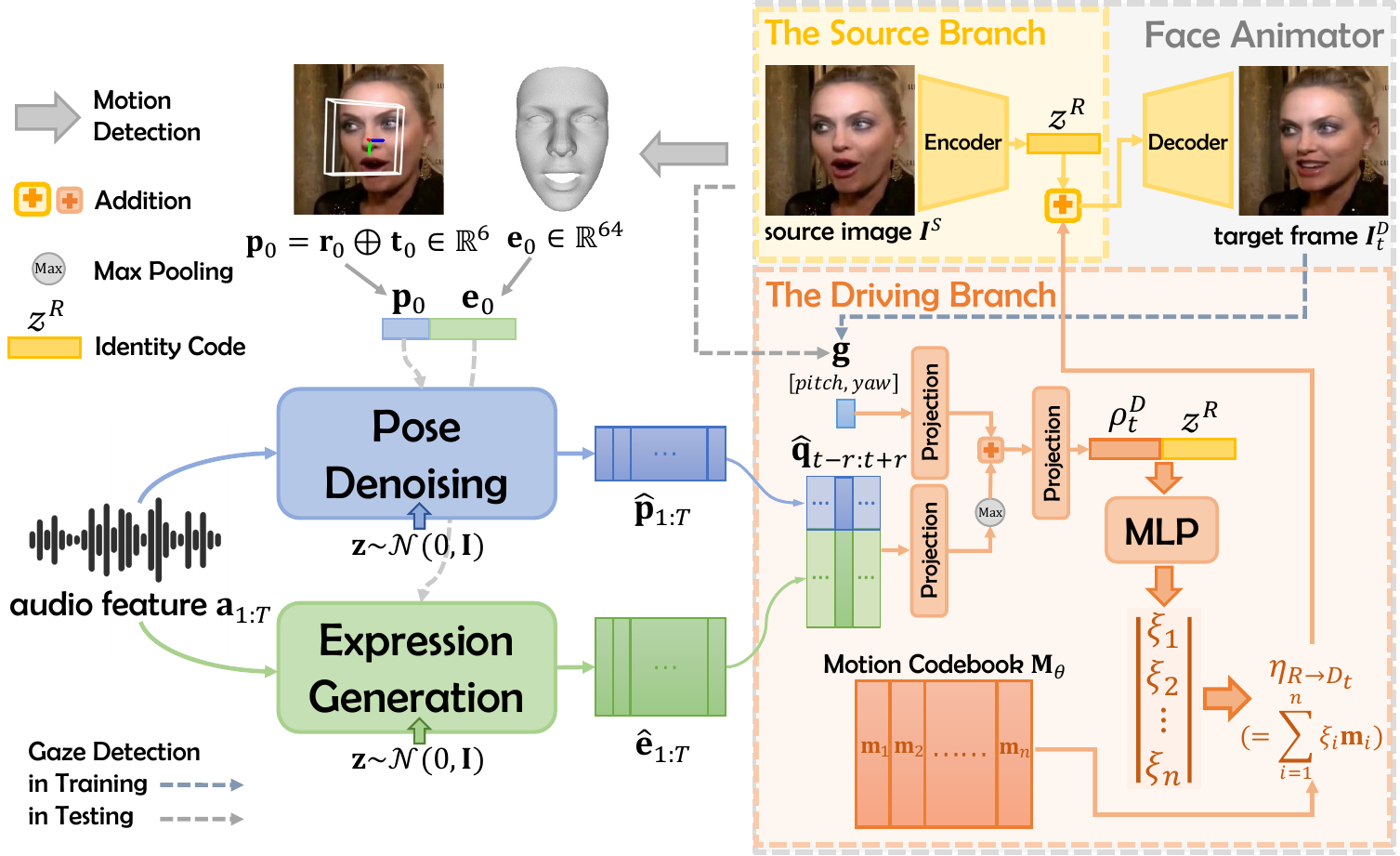}
    \caption{Illustration of our proposed \textbf{GoHD}, which is a highly disentangled and controllable taking face generation framework as described at the beginning of Section 3. }
    \label{fig:pipeline}
\end{figure}

The whole pipeline of our method is shown in Fig. \ref{fig:pipeline}. Given an audio clip with $T$ frames of mel-spectrogram ($\mathbf{a}_{1:T}$) and an input source image $\boldsymbol{I}^S$, denoting its original pose and expression coefficients as $\mathbf{p}_0$ and $\mathbf{e}_0$ respectively, with a driving gaze direction $\mathbf g$ (detected from $\boldsymbol{I}^S$ or personally defined), the sequential talking face frames $\hat{\boldsymbol{I}}^D_{1:T}$ are generated as follows:\\ 
\noi{1) Diffused Head Poses.} A rhythmic head pose sequence $\hat{\mathbf p}_{1:T}$ is synthesized through a probabilistic diffusion model conditioned on the input audio frames $\mathbf{a}_{1:T}$ and the original parameter $\mathbf{p}_0$.\\
\noi{2) Audio-to-expression Prediction.} A sequence of expression coefficients, denoted as $\hat{\mathbf e}_{1:T}$, is obtained by a predictor trained in two stages, integrating an MLP-based (Multilayer Perceptron) distillation network and a generative LSTM model, from the given audio segment $\mathbf{a}_{1:T}$ and the original expression parameter $\mathbf{e}_0$. \\
\noi{3) Gaze-oriented Face Animation.} Given the predicted motion descriptors $\hat{\mathbf p}_{1:T}$ and $\hat{\mathbf e}_{1:T}$, and the predetermined gaze orientation $\mathbf g$, the input source image $\boldsymbol{I}^S$ can be animated to $\hat{\boldsymbol{I}}^D_{1:T}$ frame by frame through an animation module involving latent space navigation to achieve robust facial motion transformations. 

\begin{figure}
    \centering
    \includegraphics[width=\linewidth]{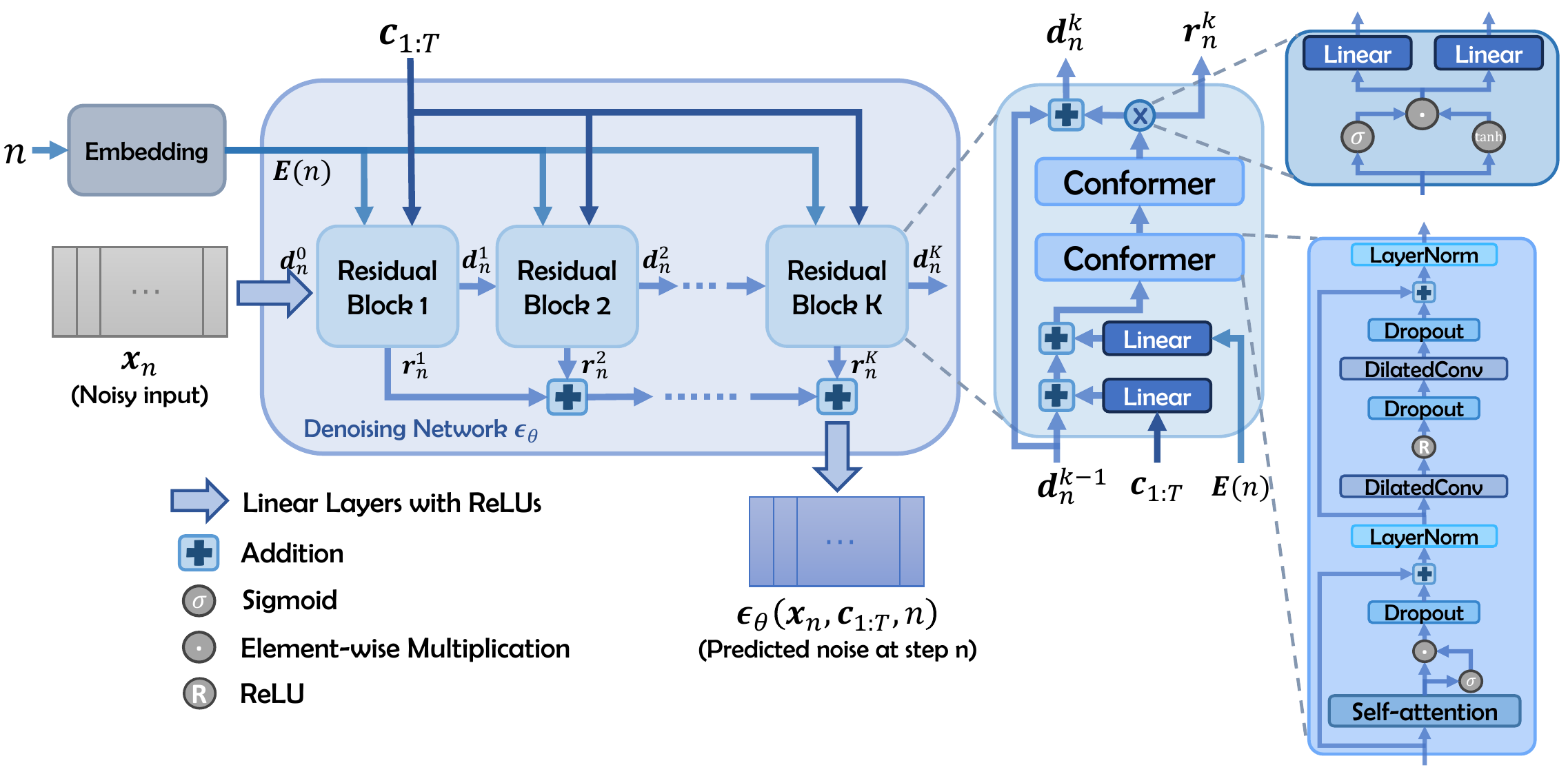}
    \caption{Demonstration of the residual denoising network architecture in the diffusion model for head pose estimation.}
    \label{fig:pose_gen}
    \vspace{-0.3cm}
\end{figure}

\subsection{Diffusion-based Head Pose Generator}
\noi{Diffusion Model.} 
To ensure that the generated head motions exhibit diversity while maintaining a sense of rhythmicity, we design a conditional diffusion model to synthesis head pose coefficients $\hat{\mathbf{p}}_{1:T}$ corresponding to the input audio feature $\mathbf{a}_{1:T}$. The diffusion process at step $n\in\{1,2,...,\mathbf{N}\}$ is defined as follows:
\begin{equation}
    \label{eq:diffusion} q(\boldsymbol{x}_n | \boldsymbol{x}_{n-1})=\mathcal{N}(\boldsymbol{x}_n; \sqrt{\alpha_n}\boldsymbol{x}_{n-1}, \beta_n\mathbf{I}), 
\end{equation}
where $\alpha_n=1-\beta_n(0<\beta_n<1)$ \shortcite{ho2020denoising}, so that $\{\beta_n\}^\mathbf{N}_{n=1}$ completely defines the diffusion process by sampling $\boldsymbol{x}_n=\sqrt{1-\beta_n}\boldsymbol{x}_{n-1}+\sqrt{\beta_n}\boldsymbol\epsilon$ with $\boldsymbol \epsilon \sim \mathcal{N}(\mathbf{0}, \mathbf{I})$. In our context, we designate the original variable $\boldsymbol{x}_0$ as the residual sequence of the ground-truth pose coefficients $\Delta \mathbf{p}_{1:T}=\mathbf{p}_{1:T} - \mathbf{p}_0 \in \mathbb{R}^{T \times 6}$ to generate more natural and continuous motion over the first pose of the sequence.

As formulated by DDPM \shortcite{ho2020denoising}, the network only needs to predict the added noise $\boldsymbol{\epsilon}$, thus the loss function can be constructed as:
\begin{equation}
    \begin{split}
    \label{eq:diff_loss} \mathcal{L}(\theta|\mathcal{D})=\mathbb{E}_{\boldsymbol{x}_0,n,\boldsymbol{\epsilon}} \big[ \Vert \boldsymbol{\epsilon}-\boldsymbol{\epsilon}_\theta(\sqrt{\bar{\alpha}_n} \boldsymbol{x}_0+\sqrt{\bar{\beta}_n} \boldsymbol{\epsilon}, \boldsymbol{c}, n) \Vert^2 \big],  
    \end{split}
\end{equation}
where $\boldsymbol{x}_0$ is uniformly sampled from the training data $\mathcal{D}$, $\bar{\alpha}_n$ and $\bar{\beta}_n$ are constants determined by $\{\beta_n\}^\mathbf{N}_{n=1}$, and $\boldsymbol{\epsilon}_\theta$ is the conditional noise prediction network with learnable parameters. In our case, the condition variable $\boldsymbol{c}$ can either be the input audio feature $\mathbf{a}_{1:T}$ or its combination with the initial pose coefficient $\mathbf{p}_0$.

\noi{Network Architecture.}
The architecture of our denoising network is shown in Fig. \ref{fig:pose_gen}. Inspired by \cite{kong2021diffwave}, we implement the network with a series of conditional residual blocks for generating audio-aware residual pose sequences. Within each block, we stack two conformers where attention modules are incorporated into dilated convolutions to effectively assimilate information over extended time scales. 

\noi{Head Pose Synthesis.}
To enhance the realism of the synthesized results, we incorporate classifier-free guidance \cite{ho2021classifierfree} to partially condition the reverse diffusion process on the initial pose ${\mathbf p}_0$. Given an input reference pose $\mathbf{p}_0$, we define the source-referred conditioning $\boldsymbol{c}_{1:T}$ with $\boldsymbol{c}_t=\mathbf{a}_t\oplus\mathbf{p}_0$, where $\mathbf{a}_t$ is the audio feature at the \emph{t}-th frame. After separately training a $\mathbf{p}_0$-conditional model $\boldsymbol{\epsilon}_\theta(\boldsymbol{x}_{n}, \boldsymbol{c}_{1:T}, n)$ and a $\mathbf{p}_0$-unconditional model $\boldsymbol{\epsilon}_\theta(\boldsymbol{x}_{n}, \mathbf{a}_{1:T}, n)$, the classifier-free guidance can be achieved by combining the prediction of both models:
\begin{equation}
    \label{eq:classifier-free} 
    \begin{split}
    \boldsymbol{\epsilon}_\theta^\gamma(\boldsymbol{x}_{n}, \boldsymbol{c}_{1:T}, n)=\gamma \boldsymbol{\epsilon}_\theta(\boldsymbol{x}_{n}, \boldsymbol{c}_{1:T}, n) + (1-\gamma)\boldsymbol{\epsilon}_\theta(\boldsymbol{x}_{n}, \mathbf{a}_{1:T}, n),
    \end{split}
\end{equation}
where the coefficient $\gamma(0<\gamma<1)$ can be adjusted to control the influence of $\mathbf{p}_0$ 
, and the synthesized head pose sequence can be inferred by: 
\begin{equation}
    \label{eq:pose_synthesis}\hat{\mathbf{p}}_{1:T} = \mathbf{p}_0 + \hat{\Delta}\mathbf{p}_{1:T} = \mathbf{p}_0 + \hat{\boldsymbol{x}}_0.
\end{equation}
where $\hat{\boldsymbol{x}}_0$ is the reverse sampling result given the predicted noise.

\begin{figure}
    \centering
    \includegraphics[width=0.9\linewidth]{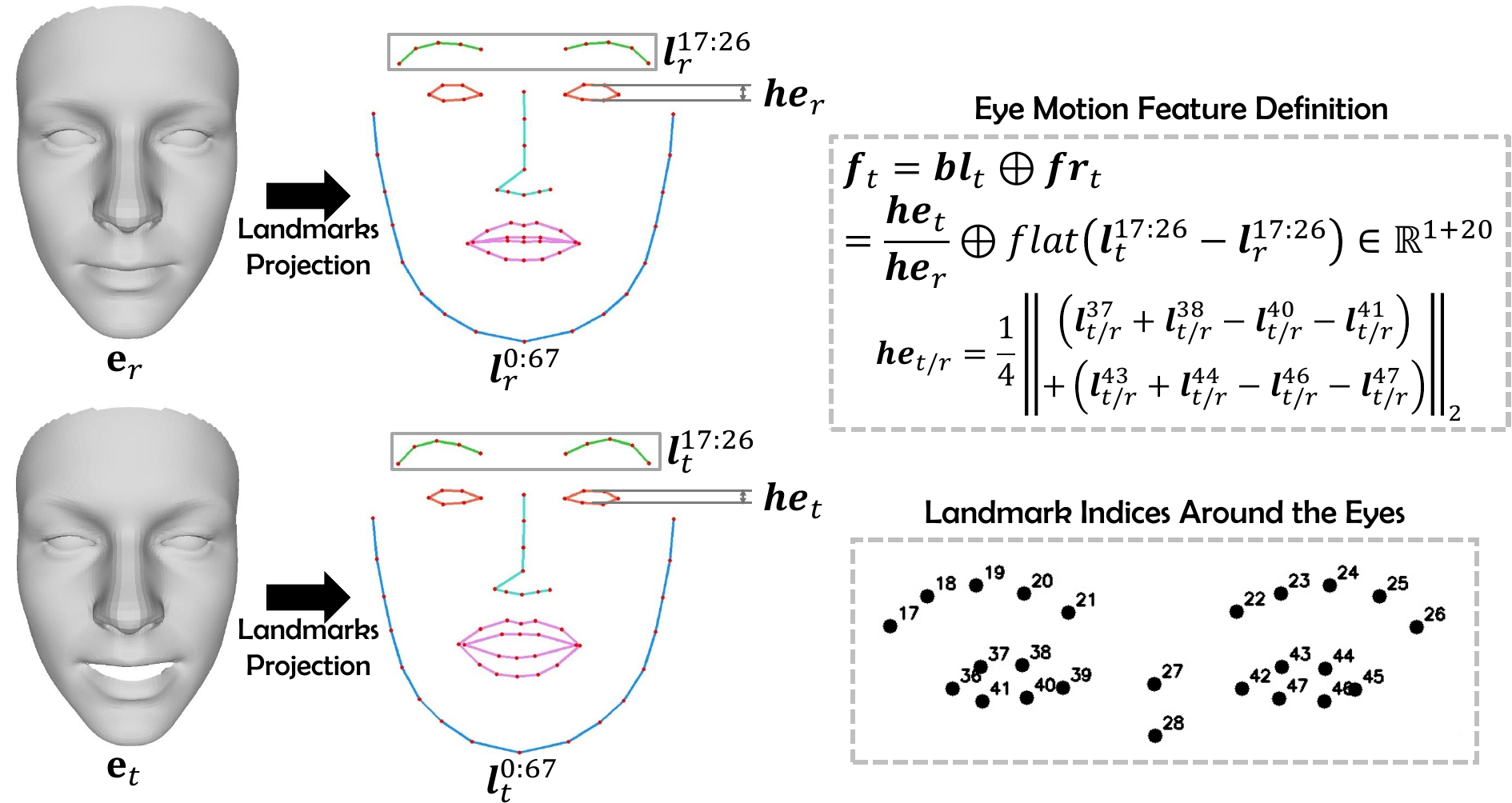}
    \caption{Definition of the eye motion feature, where $\boldsymbol{bl}_{t}$ represents the eye-blinking ratio of the \emph{t}-th frame, with $\boldsymbol{he}_{t/r} \in \mathbb{R}$ denoting the average heights of eyes. $\boldsymbol{fr}_t \in \mathbb{R}^{20}$ symbolizes the corresponding brow displacements, and $flat$ means the operation of flattening. The landmark indices and calculation for ${\boldsymbol{he}}_{t/r}$ are illustrated on the right side.} 
    \label{fig:eye_motion}
    \vspace{-0.3cm}
\end{figure}

\subsection{Expression Predictor Trained in Two Stages}
This module includes frame-wise distillation for audio-synchronized mouth shapes and temporal prediction for spontaneous eye motions. To facilitate the disentanglement of various facial actions from the overall expression coefficients, we pre-define handcrafted eye motion features as control signals for the first stage and as the generation goal of the second stage.

\noi{Handcrafted Eye Motion Features.}
Given the expression coefficients $\mathbf{e}_t$ of the \emph{t}-th frame, the facial mesh can be reconstructed by setting all other coefficients (poses and identity) to zero. We then extract the 68 facial landmarks $\boldsymbol{l}^{0:67}_t$ from the above mesh. Similarly, a reference set of landmarks $\boldsymbol{l}^{0:67}_r$ is obtained from a neutral facial mesh with ``mean expression'' $\mathbf{e}_r$, where all coefficients of the expression basis are set to zero. Considering eye blinks and brow frowns, we define the eye motion feature $\boldsymbol{f}_t \in \mathbb{R}^{21}$ in Fig. \ref{fig:eye_motion}.


\noi{Stage 1: Audio-to-lip Distillation.}
The audio-to-lip mapping poses a one-to-one problem due to the strong connection between mouth shape and pronunciation. To ensure that the network specifically learns the correlation between audio and lip motions in the first stage, we incorporate the ground-truth handcrafted eye motion features $\boldsymbol{f}_{1:T}$ as additional input signals along with the audio $\mathbf{a}_{1:T}$ and the initial expression coefficients $\mathbf{e}_0$ for regressing the overall expressions. The mapping of each frame can be written as:
\begin{equation}
    \label{eq:mapping} \tilde{\mathbf{e}}_t=\mathbf{MLP}_\theta(\mathbf{\Phi}_a(\mathbf{a}_t)\oplus\mathbf{e}_0\oplus\boldsymbol{f}_t),
\end{equation}
where $\mathbf{\Phi}_a$ is an audio encoder that embeds the input audio feature to a latent space and $\mathbf{MLP}_\theta$ denotes a multilayer perception. Notably, we distill the resynchronized results from a pre-trained lip expert \cite{wav2lip} ($\mathcal{L}_{distill}$) to inherit its lip-audio alignment capability learned on sufficient sample pairs, thereby compensating for our limited dataset and reducing the risk of under-fitting. 

\begin{figure}
    \centering
    \includegraphics[width=\linewidth]{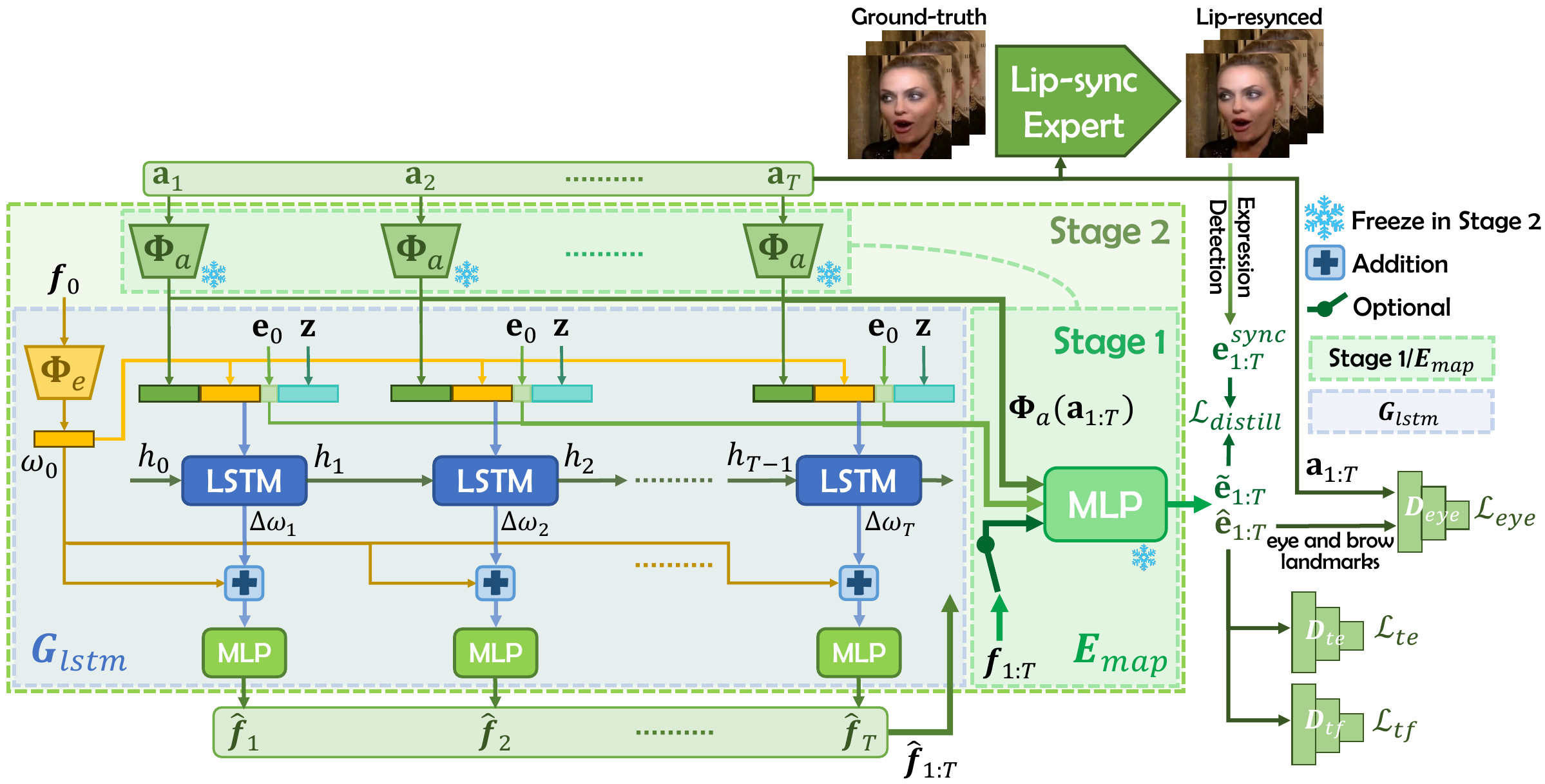}
    \caption{The expression predictor trained in two stages.}
    \label{fig:exp_gen}
    \vspace{-0.3cm}
\end{figure}

\noi{Stage 2: Eye Motion Generation.}
After learning synchronized audio-to-lip mapping, the second stage focuses on addressing the more complex mapping between audio and eye motions and is trained with the learned weights in stage 1 frozen. To tackle this generation problem containing more temporal dynamics, we employ the LSTM architecture. This choice over transformer models is deliberate, as LSTMs are known for their robustness in handling longer sequences during inference, ensuring effective modeling of information dependencies across various time scales. As is depicted in Fig. \ref{fig:exp_gen}, taking the sequence of audio features $\mathbf{a}_{1:T}$, the initial expression coefficients $\mathbf{e}_0$, and the corresponding eye motion feature $\boldsymbol{f}_0$ as input, we first encode $\mathbf{a}_{1:T}$ through the audio encoder $\mathbf{\Phi}_a$ pre-trained in the first stage. Along with $\mathbf{e}_0$ and the encoded eye motion features $\omega_0 = \mathbf{\Phi}_e(\boldsymbol{f}_0)$, the sequential procedure can be described as follows:
\begin{align}
    \label{eq:lstm1} (h_t, \Delta\omega_t)&=\mathbf{LSTM}_\theta(h_{t-1}, \mathbf{\Phi}_a(\mathbf{a}_t)\oplus\omega_0\oplus\mathbf{e}_0\oplus\mathbf{z}),\\
    \label{eq:lstm2} \hat{\boldsymbol{f}}_t&=\mathbf{MLP}_\theta(\omega_t)=\mathbf{MLP}_\theta(\omega_0+\Delta\omega_t),
\end{align}
where $h_t$ is the hidden state at time step $t$, which corresponds to the \emph{t}-th frame, while $h_0$ is a zero vector with the same shape as $\omega_0$. To encourage the network to learn multiple probabilities of generating spontaneous motions, we also concatenated a latent vector $\mathbf{z}$ in the hidden layer at each step, which is sampled from the standard multivariate Gaussian distribution. Note that we have the network predict the residuals of the embedded eye motion features for faster convergence and better generalization ability. Combining with the well-trained mapping network from the first stage, the overall estimation of audio-driven expressions is completed in a single, cohesive process:
\begin{align}
    \notag \hat{\mathbf{e}}_{1:T} &= \boldsymbol{E}_{map}(\mathbf{a}_{1:T}, \mathbf{e}_0, \hat{\boldsymbol{f}}_{1:T})\\
    \label{eq:exp} &=\boldsymbol{E}_{map}(\mathbf{a}_{1:T}, \mathbf{e}_0, \boldsymbol{G}_{lstm}(\mathbf{\Phi}_a(\mathbf{a}_{1:T}), \boldsymbol{f}_0, \mathbf{e}_0, \mathbf{z})),
\end{align}
where $\boldsymbol{E}_{map}$ and $\boldsymbol{G}_{lstm}$ are the frame-wise mapping network and the LSTM-based eye motions generator, respectively. 
We introduce three discriminators $\boldsymbol{D}_{eye}$, $\boldsymbol{D}_{te}$ and $\boldsymbol{D}_{tf}$ to help distinguish the temporal naturalness and realness of the results. Extended descriptions can be found in the \emph{supplementary material}. 
\subsection{Latent Navigable Face Animator}
Given the pose and expression coefficients $\hat{\mathbf p}_{1:T}$ and $\hat{\mathbf e}_{1:T}$ predicted from the audio, along with the reference image $\boldsymbol{I}^S$ and target gaze direction $\mathbf g$, we draw inspiration from \cite{wang2022latent} and introduce a well-designed face animator to generate the final talking portrait 
 frames $\hat{\boldsymbol{I}}^D_{1:T}$. Unlike previous methods that rely on the transformations of spatial key points \cite{wang2021facevid2vid, Siarohin_2019_NeurIPS}, our animator directly manipulates the latent space to alleviate information loss caused by using explicit structural representations and achieve better disentanglement of identity and motion. Additionally, different from \cite{wang2022latent} that requires a real video as the overall motion-driving signal, our animator makes the animation derivable through separate intermediate motion descriptors. This design choice enables explicit editing of various facial attributes and supports multi-modal driving (Fig. \ref{fig:PEGcontrol}). In training time, it animates the source image in a frame-by-frame manner by learning the motion transformation from $\boldsymbol{I}^S$ to the target \emph{t}-th frame $\boldsymbol{I}^D_t$ via the detected coefficients $\mathbf{q}_{t-r:t+r}=\mathbf{p}_{t-r:t+r}\oplus \mathbf{e}_{t-r:t+r}$, where $\boldsymbol{I}^S$ and $\boldsymbol{I}^D_t$ are two randomly selected frames of a video, and $r$ is the radius of the adjacent window for smoothing, which is achieved by a max pooling layer after several layers of projection. As shown in Fig. \ref{fig:pipeline}, we first encode the source image into a latent space to acquire an identity code $z^R$. This latent vector is then concatenated with the projected and gaze-conditioned driving feature $\rho^D_t$ to estimate the motion transfer $\eta_{R\rightarrow D_t}$ on a learnable motion codebook, which consists of a series of learnable orthogonal motion directions $\mathbf{M}_\theta=\{\mathbf{m}_1, ... \mathbf{m}_n\}$ to represent any latent navigation. By jointly learning the magnitude $\xi_i$ of each direction $\mathbf{m}_i$, the latent navigation can be linearly calculated as follows:
\begin{align}
    \label{eq:lsn} \eta_{R\rightarrow D_t}=\sum_{i\in[1,n]}\xi_i\mathbf{m}_i,
\end{align}
where $\xi_{1:n}=\mathbf{MLP}_\theta(\rho^D_t\oplus z^R)$. 
Afterward, the target latent representation can be obtained by simple addition: $z^D_t=z^R+\eta_{R\rightarrow D_t}$, from which the target frame $\boldsymbol{I}^D_t$ will be generated through a decoder. Notice that, during training, the driving gaze directions are directly inherited from the driving frames, then the module's gaze orientation ability can be optimized through a simple gaze loss. 
During inference, the driving gaze directions can be set to any reasonable pitch and yaw angles to achieve effective gaze manipulation or rectify potentially unnatural looking directions. For simplicity, we set them to the original gaze directions derived from $\boldsymbol{I}^S$s in most of our experiments.

\section{Experiments}
\begin{figure*}[h]
    \centering
    \includegraphics[width=\linewidth]{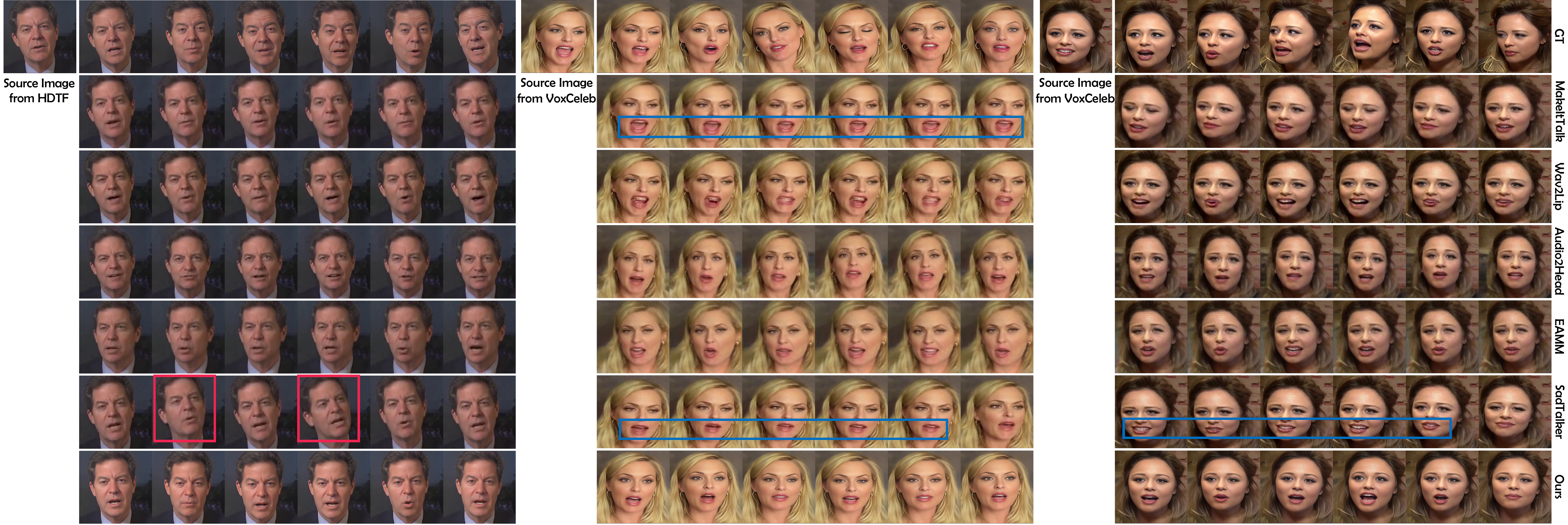}
    \caption{Qualitative comparison on the two datasets. Apart from accurate lip synchronization, our method presents the best generalization capability on animating extravagant input expressions and stability in pose generation.}
    \label{fig:compare}
    \vspace{-0.3cm}
\end{figure*}
\subsection{Experimental Settings}
\noi{Datasets.} We leverage a subset of the VoxCeleb dataset \shortcite{nagrani2017voxceleb} as the training set for our face animator, and a portion of the HDTF dataset \shortcite{zhang2021flow} for the generation of motion descriptors. Most of the testing is also conducted on hundreds of unseen videos from these two datasets. 

\begin{table}[t]
  \centering
  \resizebox{\linewidth}{!}{
    \renewcommand\arraystretch{1.1}
    \begin{tabular}{c|ccc|ccc}
    \toprule
    \multirow{2}[4]{*}{Method} & \multicolumn{3}{c|}{HDTF} & \multicolumn{3}{c}{VoxCeleb} \\
\cmidrule{2-7}          & LSE-C $\uparrow$ & LSE-D $\downarrow$ & FID $\downarrow$   & LSE-C $\uparrow$ & LSE-D $\downarrow$ & FID $\downarrow$ \\
    \cline{1-7}
    \cline{1-7}
    \cline{1-7}
    MakeItTalk \shortcite{zhou2020makelttalk} & 5.58  & 9.85  & 34.28 & 4.40  & 10.42 & 61.98  \\
    Wav2Lip \shortcite{wav2lip} & \textbf{10.04} & \textbf{5.93} & 34.40 & \textbf{9.32} & \textbf{6.15} & 67.75 \\
    Audio2Head \shortcite{wang2021audio2head} & 7.97  & \underline{7.30}  & 34.31 & 5.79  & 8.61  & 79.05  \\
    EAMM \shortcite{eamm} & 5.45  & 9.57  & 57.34 & 4.74  & 9.61  & 85.39 \\
    SadTalker \shortcite{Zhang_2023_CVPR} & 7.60  & 7.70  & 36.91 & 6.99  & 7.75  & 60.75 \\
    Ours  & \underline{8.13}  & 7.78  & \textbf{30.40} & \underline{7.20}  & \underline{7.70}  & \textbf{58.11} \\
    \cline{1-7}
    Ground Truth    & 8.97  & 6.67  &   -   & 7.51  & 7.42  &   -  \\
    \bottomrule
    \end{tabular}}%
  \caption{Quantitative comparisons for lip synchronization and video quality, with the best results highlighted in \textbf{bold}. Given that scores obtained through the pre-trained SyncNet model \cite{wav2lip} may not serve as a definitive criterion, we additionally \underline{underline} values closest to the ground truth, providing a more authoritative reference.}
   \label{tab:videoquality}%
  \vspace{-0.1cm}
\end{table}%

\noi{Comparison Methods.}
We conduct a comprehensive evaluation of our method by comparing it with various advanced audio-only driven methods, including Wav2Lip \cite{wav2lip}, MakeItTalk \cite{zhou2020makelttalk}, Audio2Head \cite{wang2021audio2head}, EAMM \cite{eamm}, and SadTalker \cite{Zhang_2023_CVPR}. Our evaluation covers lip synchronization and video quality for all the mentioned approaches. Additionally, we assess the data structural match and naturalness of other spontaneous motions when compared to SadTalker and Audio2Head. 

\begin{table}[t]
  \centering
  \resizebox{\linewidth}{!}{
    \renewcommand\arraystretch{1.1}
    \begin{tabular}{c|cccc|cccc}
    \toprule
    \multirow{2}[4]{*}{Method} & \multicolumn{4}{c|}{HDTF}                      & \multicolumn{4}{c}{VoxCeleb} \\
\cmidrule{2-9}          & $\text{Var}_p^{\times 10^3}$ & $\text{SSIM}_p$ $\uparrow$ & $\text{Var}_e^{\times 10^2}$ & $\text{SSIM}_e$ $\uparrow$ & $\text{Var}_p^{\times 10^3}$ & $\text{SSIM}_p$ $\uparrow$ & $\text{Var}_e^{\times 10}$ & $\text{SSIM}_e$ $\uparrow$ \\
    \cline{1-9}
    \cline{1-9}
    \cline{1-9}
    Audio2Head \shortcite{wang2021audio2head} & 2.399 & 0.972 & 3.066 & 0.819 & 1.992 & 0.984 & 0.618 & 0.754 \\
    SadTalker \shortcite{Zhang_2023_CVPR} & \underline{2.473} & 0.985 & 2.206 & 0.904 & 1.896 & \textbf{0.987} & 0.182 & 0.854 \\
    Ours  & 2.411 & \textbf{0.996} & \underline{5.374} & \textbf{0.915} & \underline{2.314} & \textbf{0.987} & \underline{0.865} & \textbf{0.872} \\
    \cline{1-9}
    Ground Truth    & 4.315 &   -   & 9.778 &   -   & 8.746 &   -   & 1.585 &  - \\
    \bottomrule
    \end{tabular}}%
  \caption{Quantitative comparisons for spontaneous motions, with the best SSIM scores highlighted in \textbf{bold}. As for the variances, we \underline{underline} the values that are closest to the ground truth to indicate a statistical match.}
  \label{tab:motions}%
  \vspace{-0.12cm}
\end{table}%


\noi{Evaluation Metrics.}
We use Frechet Inception Distance (FID) \cite{FID} to evaluate image quality. For lip synchronization, we adopt methods from previous works \cite{ Zhang_2023_CVPR, yu2023thpad} and utilize the pre-trained SyncNet \cite{wav2lip} for confidence (LSE-C) and distance (LSE-D) evaluations of lip motions. Using the 2D landmarks derived from the detected expression coefficients, we compute the structural similarity ($\text{SSIM}_{e}$) and average variance ($\text{Var}_{e}$) on the eyes and brows landmarks sequences to assess the naturalness of eye motions. On the other hand, to evaluate poses, we employ a pre-trained pose detection model \cite{ALGABRI2024122293HPE} to obtain the pose sequences of the generated videos. We then calculate the structural similarities between these sequences ($\text{SSIM}_{p}$) and the average variance of their corresponding feature vectors ($\text{Var}_{p}$) to indicate their statistical match with real data.

\subsection{Comparison with State-of-the-art Methods}
\noi{Quantitative Comparison.} Quantitative evaluations for lip synchronization and video quality are reported in Table \ref{tab:videoquality}. According to the FID, our approach demonstrates an overall improvement in the realism of generated videos. In terms of lip-sync performance, Wav2Lip unquestionably achieves the best results, surpassing even the ground truth, because it directly trains with the SyncNet model used for evaluation. Consequently, we interpret scores closer to the ground truth as indicating a relatively better ability to produce realistic mouth movements. In this context, our method exhibits better performance than SadTalker. Meanwhile, Audio2Head presents smaller lip motion distances on the HDTF dataset, likely due to the overlap between our testing set and its training set. Furthermore, Table \ref{tab:motions} illustrates assessments for spontaneous motions. Two representative methods (Audio2Head and SadTalker) are included in this comparison. Audio2Head exhibits high diversity in generated poses and expressions but suffers from significant misalignment with real data, especially in expressions. In contrast, SadTalker demonstrates good structural similarity with the ground truth, albeit with lower diversity, especially in eye motions, as it only considers controllable blinks in expression generation. Our GoHD achieves a balance between data diversity and realism, presenting a comprehensive advancement in poses and eye motion generation.


\noi{Qualitative Comparison.} Fig. \ref{fig:compare} shows visual comparisons of three examples from the HDTF and the VoxCeleb dataset. Audio2Head \cite{wang2021audio2head} and EAMM \cite{eamm} both rely on the animation framework of FOMM \cite{Siarohin_2019_NeurIPS}, exhibiting severe face distortions and struggling to preserve identity. MakeItTalk \cite{zhou2020makelttalk} performs poorly in lip synchronization, while Wav2Lip \cite{wav2lip} suffers from artifacts in the lip region, especially when handling substantial variations in mouth shape. SadTalker demonstrates relatively high visual quality but occasionally produces unnatural and upward-tilted head poses. As shown on the middle and the right, it encounters incomplete motion disentanglement and has difficulty animating faces with exaggerated lip morphology. In general, aside from accurate lip synchronization, our method demonstrates superior generalization capability in animating extravagant input expressions and stability in pose generation. 


\begin{figure}[t]
    \centering
    \includegraphics[width=\linewidth]{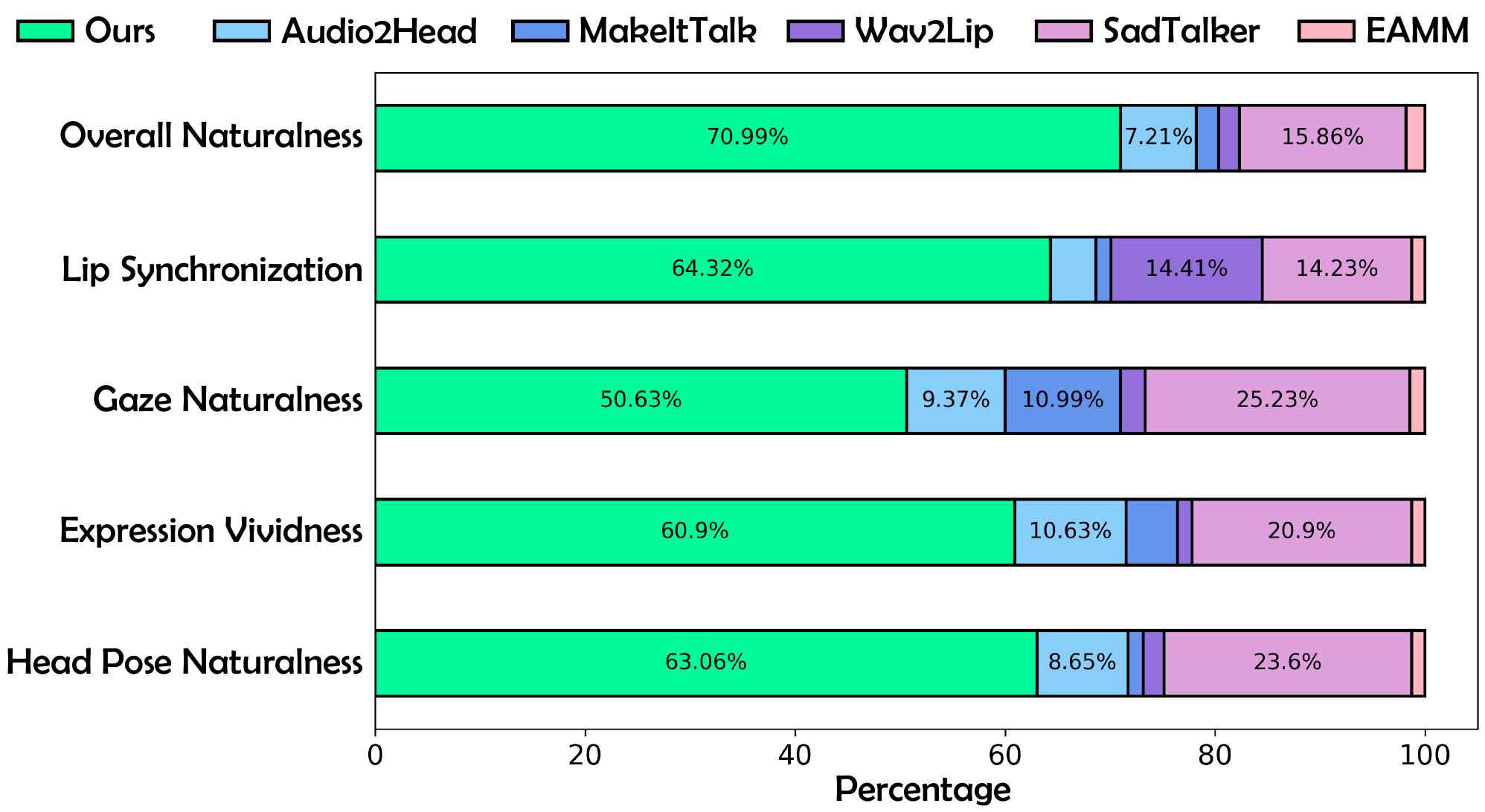}
    \caption{The result of user study.}
    \label{fig:user}
    \vspace{-0.5cm}
\end{figure}

\noi{User Study.}
We conduct a user study to evaluate the overall performance of our method against various competitors. We randomly select 30 test examples and invite 37 volunteers to assess each example in terms of head pose naturalness, expression vividness (with a focus on eye motions like blinks and frowns), gaze naturalness, lip synchronization, and overall naturalness. With a total of $30 \times 37 = 1,110$ responses for each attribute, the support percentages for each method are depicted in Fig. \ref{fig:user}. Notably, our method outperforms all others on comprehensive aspects, receiving 70.99\% of the responses for overall naturalness. 

\subsection{Validation Experiments}
\begin{figure}
    \centering
    \includegraphics[width=\linewidth]{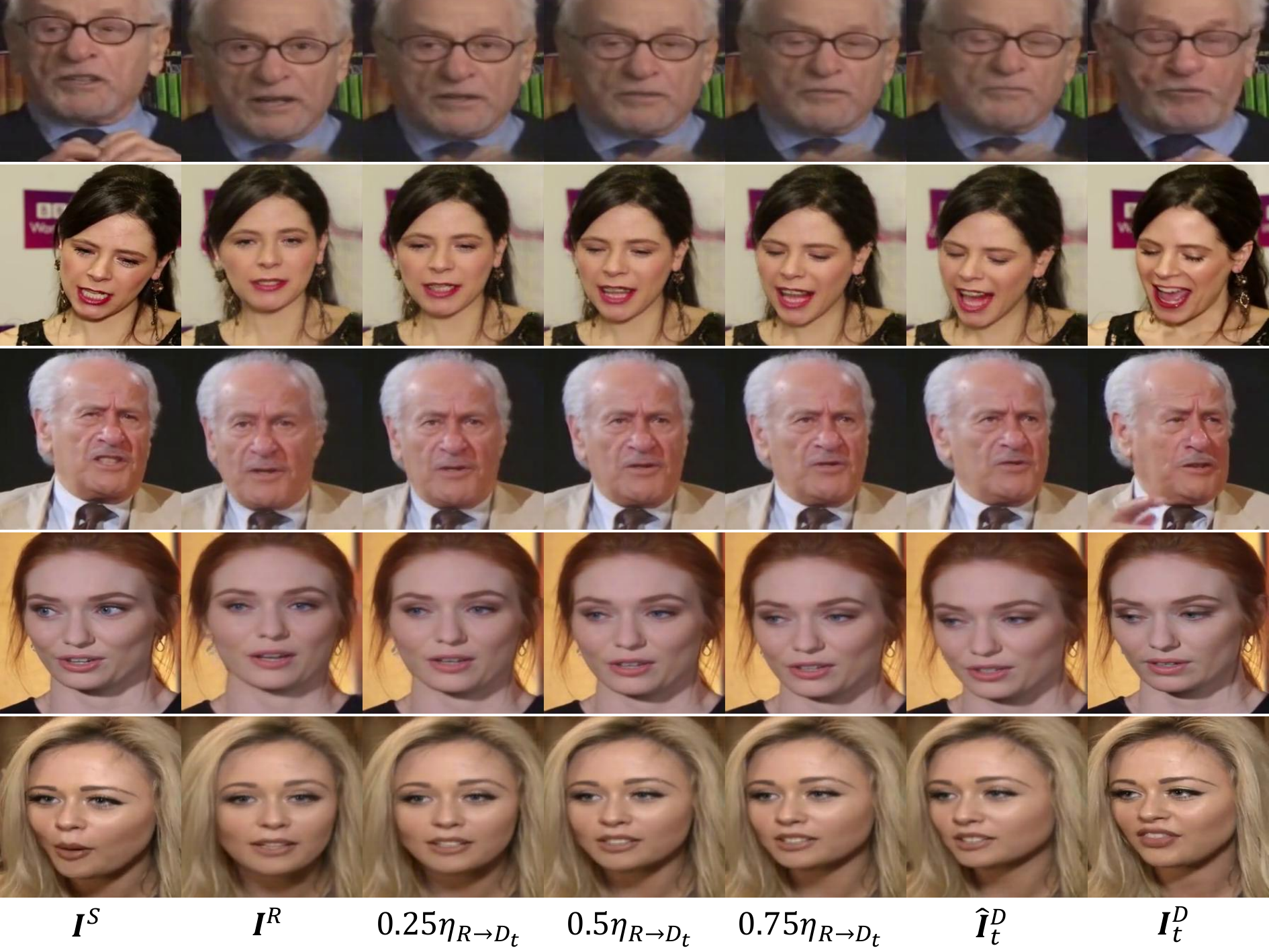}
    \caption{Visualization of motion interpolation. We can see a gradual transfer in all motions as the latent navigation vector $\eta_{R\rightarrow D_t}$ linearly increases, representing effective motion and identity disentanglement.}
    \label{fig:interpolation}
\end{figure}

\noi{Motion Interpolation.}
We provide visualizations for motion interpolation of our face animator to showcase its robustness in motion editing. The reference images $\boldsymbol{I}^R$s, decoded from the reference latent representations $z^R$s, consistently exhibit a frontal pose and mean expression, demonstrating the effective disentanglement of motion from identity. In Fig. \ref{fig:interpolation}, as the coefficient $\lambda$ of the latent navigation vector $\eta_{R\rightarrow D_t}$ linearly increases, the final image derived from $z^R+\lambda \eta_{R\rightarrow D_t}$ (denoted by $\lambda \eta_{R\rightarrow D_t}, \lambda \in{0.25,0.5,0.75}$ in the figure) gradually transfers in all motions, until $\lambda=1$ to reach the target ones, indicating the effectiveness and versatility of our method in controllable motion transformation.

\noi{Gaze Orientation.}
To validate the gaze control capability of our face animator, we set the yaw angle in the driving gaze direction to three different values ($0$, $0.3\pi$, $-0.3\pi$), corresponding to looking forward, left, and right, respectively. Using the same audio clip, the results for two identities are shown in Fig. \ref{fig:gaze}, demonstrating effective manipulation on gaze orientation across various input portraits.
\begin{figure}[t]
    \centering
    \includegraphics[width=\linewidth]{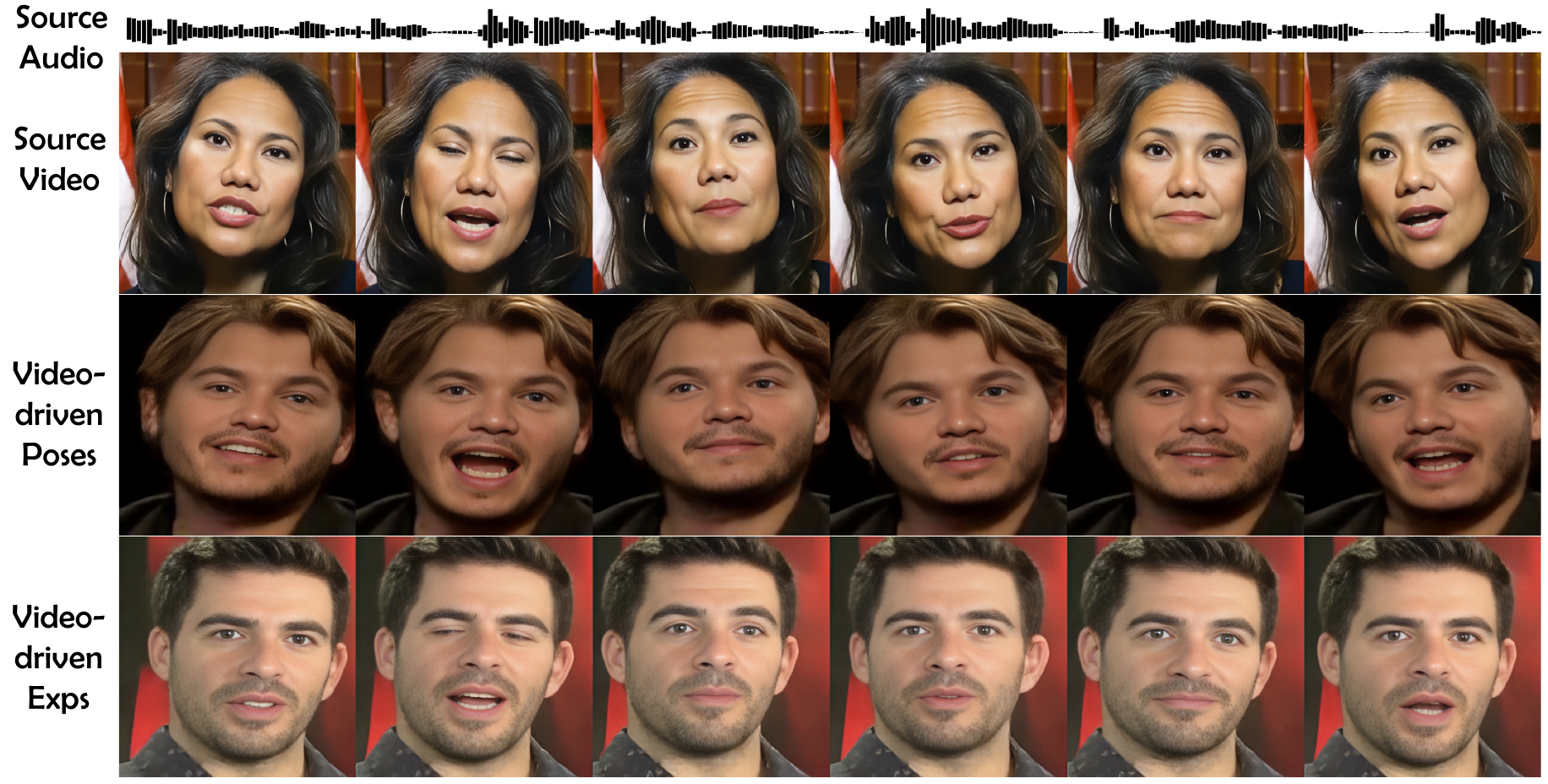}
    \caption{Demonstration of multi-modal driving results.}
    \label{fig:PEGcontrol}
    \vspace{-0.3cm}
\end{figure}

\noi{Multi-modal Driving.} 
In addition to solely relying on audio-driven generation, our approach allows for the extraction of intermediate motion descriptors from a source video, enabling multimodal-driven animation. As illustrated in Fig. \ref{fig:PEGcontrol}, the "Video-driven Exps" scenario involves deriving $\mathbf e_{1:T}$ from the source video and the $\mathbf p_{1:T}$ from the source audio, and vice versa. The video-driven signals in the results align with those of the source video, while allowing variations in the audio-driven component. These results demonstrate the effectiveness of our approach in achieving disentangled control over facial animations, with promising implications for multi-modal applications.

\subsection{Ablation Study}

\begin{table}[t]
  \centering
    \renewcommand\arraystretch{1.1}
    \begin{tabular}{l|ccc}
    \toprule
    Strategy & MLD $\downarrow$   & $\text{SSIM}_e$ $\uparrow$\\
    \cline{1-3}
    \cline{1-3}
    \cline{1-3}
    w/o \textbf{Stage 2}\&$\boldsymbol{f}_{1:T}$ & \textbf{1.785} & 0.813 \\
    w/o \textbf{Stage 1} & 1.931 & 0.836 \\
    w/o Distillation & 2.012 & 0.852 \\
    \textbf{Full}-transformer & 1.806 &  0.894 \\
    \textbf{Full}-LSTM & 1.792 &  \textbf{0.915} \\
    \bottomrule
    \end{tabular}%
  \caption{Ablation results for the two-stage strategy in expression prediction. The best results are highlighted in \textbf{bold}.}
  \label{tab:twostage_ab}%
  \vspace{-0.2cm}
\end{table}%


  

\noi{Two-stage Strategy.}
To verify the effectiveness of our carefully designed two-stage expression predictor, we conduct ablation studies on 100 videos of the HDTF dataset with the following variants: 1) w/o \textbf{Stage 2}\&$\boldsymbol{f}_{1:T}$: Produce the expression coefficients in a regressive manner by only employing the mapping network in Stage 1 without inputting eye motions features. 2) w/o \textbf{Stage 1}: Generate expressions directly through Stage 2 without pre-training the Stage 1 network. 3) w/o Distillation: Use the ground-truth lip motions as the training target in Stage 1 instead of distilling from the lip expert. 4) \textbf{Full}-transformer: Our full training strategy with $\boldsymbol{G}_{lstm}$ replaced by a transformer model. 5) \textbf{Full}-LSTM: Our full training strategy.

We compute the average mouth landmark distances (MLD) and eye motion structural similarities ($\text{SSIM}_e$) for the generated expression coefficients to evaluate each design choice in lip synchronization and eye motion generation. The numerical results are reported in Table \ref{tab:twostage_ab}. The \textbf{Full}- strategies demonstrate enhanced alignment of mouth shapes compared to the variant w/o \textbf{Stage 1}. This underscores the significant role played by the pre-trained first stage in learning lip motions synchronized with audio, where the distillation approach is also indispensable. Despite achieving the best performance in lip synchronization, employing only a mapping network to predict expressions (w/o \textbf{Stage 2}\&$\boldsymbol{f}_{1:T}$) faces challenges in producing realistic eye motions and leads to poor $\text{SSIM}_e$ score. In contrast, our \textbf{Full}- models simultaneously achieve higher lip-sync quality and naturalness in the results. Notably, the LSTM-based architecture surpasses the transformer-based one due to its ability to effectively model dependencies between neighboring frames, contributing to the overall enhanced performance in lip-sync generation by enabling more accurate sequential prediction. 

\section{Conclusion}
In this work, we introduce \textbf{GoHD}, a novel and robust framework for generating realistic audio-driven talking faces. Beyond pose and expression coefficients, we incorporate gaze direction as an additional driving condition for gaze-oriented animation. We employ a conformer-structured conditional diffusion model to synthesize rhythmic head poses. For audio-driven expression generation, we devise a predictor trained in a two-stage manner that separates frame-wise and frequent lip motions from other temporally dependent but less audio-related movements. Moreover, a latent navigable animation module is proposed for gaze-oriented and robust motion transformation. Experimental results illustrate the superiority of our GoHD to produce high-quality talking videos for any subject. 

\section*{Acknowledgments}
This work was partially funded by the National Natural Science Foundation of China (62102418, 61932003, 62372026, 62172415), the Beijing Science and Technology Plan Project (Z231100005923033), and the Excellent Youth Program of State Key Laboratory of Multimodal Artificial Intelligence Systems (MAIS2024312).


 


\bibliography{aaai25}

\begin{thebibliography}{53}
\providecommand{\natexlab}[1]{#1}

\bibitem[{Abdelrahman et~al.(2022)Abdelrahman, Hempel, Khalifa, and Al-Hamadi}]{Abdelrahman2022L2CSNetFG}
Abdelrahman, A.~A.; Hempel, T.; Khalifa, A.; and Al-Hamadi, A. 2022.
\newblock L2CS-Net: Fine-Grained Gaze Estimation in Unconstrained Environments.
\newblock \emph{ArXiv}, abs/2203.03339.

\bibitem[{Alexanderson et~al.(2023)Alexanderson, Nagy, Beskow, and Henter}]{alexanderson2023listen}
Alexanderson, S.; Nagy, R.; Beskow, J.; and Henter, G.~E. 2023.
\newblock Listen, Denoise, Action! Audio-Driven Motion Synthesis with Diffusion Models.
\newblock \emph{{ACM} {T}rans. on {G}raphics (TOG)}, 42(4): 44:1--44:20.

\bibitem[{Algabri, Shin, and Lee(2024)}]{ALGABRI2024122293HPE}
Algabri, R.; Shin, H.; and Lee, S. 2024.
\newblock Real-time 6DoF full-range markerless head pose estimation.
\newblock \emph{Expert Systems with Applications}, 239: 122293.

\bibitem[{Chen et~al.(2019)Chen, Maddox, Duan, and Xu}]{chen2019hierarchical}
Chen, L.; Maddox, R.~K.; Duan, Z.; and Xu, C. 2019.
\newblock Hierarchical cross-modal talking face generation with dynamic pixel-wise loss.
\newblock In \emph{Proc. of the {IEEE} {C}onf. on {C}omputer {V}ision and {P}attern {R}ecognition (CVPR)}, 7832--7841.

\bibitem[{Chung et~al.(2017)Chung, Jamaludin, Zisserman et~al.}]{chung2017you}
Chung, J.; Jamaludin, A.; Zisserman, A.; et~al. 2017.
\newblock You said that?
\newblock In \emph{{B}ritish {M}achine {V}ision {C}onference (BMVC)}. British Machine Vision Association and Society for Pattern Recognition.

\bibitem[{Deng et~al.(2019{\natexlab{a}})Deng, Guo, Xue, and Zafeiriou}]{deng2019arcface}
Deng, J.; Guo, J.; Xue, N.; and Zafeiriou, S. 2019{\natexlab{a}}.
\newblock Arcface: Additive angular margin loss for deep face recognition.
\newblock In \emph{Proceedings of the IEEE/CVF Conference on Computer Vision and Pattern Recognition}, 4690--4699.

\bibitem[{Deng et~al.(2019{\natexlab{b}})Deng, Yang, Xu, Chen, Jia, and Tong}]{deng2019accurate}
Deng, Y.; Yang, J.; Xu, S.; Chen, D.; Jia, Y.; and Tong, X. 2019{\natexlab{b}}.
\newblock Accurate 3D Face Reconstruction with Weakly-Supervised Learning: From Single Image to Image Set.
\newblock In \emph{Proc. of the {IEEE} {C}onf. on {C}omputer {V}ision and {P}attern {R}ecognition (CVPR)}.

\bibitem[{Drobyshev et~al.(2024)Drobyshev, Casademunt, Vougioukas, Landgraf, Petridis, and Pantic}]{drobyshev2024emoportraits}
Drobyshev, N.; Casademunt, A.~B.; Vougioukas, K.; Landgraf, Z.; Petridis, S.; and Pantic, M. 2024.
\newblock EMOPortraits: Emotion-enhanced Multimodal One-shot Head Avatars.
\newblock arXiv:2404.19110.

\bibitem[{Feng et~al.(2021)Feng, Feng, Black, and Bolkart}]{DECA}
Feng, Y.; Feng, H.; Black, M.~J.; and Bolkart, T. 2021.
\newblock Learning an Animatable Detailed 3D Face Model from In-the-Wild Images.
\newblock \emph{ACM Trans. Graph.}, 40(4).

\bibitem[{Gao et~al.(2023)Gao, Xu, Li, Yang, Zeng, and Qi}]{gao2023ctcnet}
Gao, G.; Xu, Z.; Li, J.; Yang, J.; Zeng, T.; and Qi, G.-J. 2023.
\newblock Ctcnet: a cnn-transformer cooperation network for face image super-resolution.
\newblock \emph{IEEE Transactions on Image Processing}.

\bibitem[{He et~al.(2023)He, Guo, Yu, Wang, Zhu, An, Li, Tan, Wang, Wu, Zhao, and Bian}]{he2023gaia}
He, T.; Guo, J.; Yu, R.; Wang, Y.; Zhu, J.; An, K.; Li, L.; Tan, X.; Wang, C.; Wu, H.; Zhao, S.; and Bian, J. 2023.
\newblock GAIA: Zero-shot Talking Avatar Generation.
\newblock In \emph{ICLR 2024}.

\bibitem[{Heusel et~al.(2017)Heusel, Ramsauer, Unterthiner, Nessler, and Hochreiter}]{FID}
Heusel, M.; Ramsauer, H.; Unterthiner, T.; Nessler, B.; and Hochreiter, S. 2017.
\newblock GANs Trained by a Two Time-Scale Update Rule Converge to a Local Nash Equilibrium.
\newblock In \emph{Proceedings of the 31st International Conference on Neural Information Processing Systems}, NIPS'17, 6629–6640. Red Hook, NY, USA: Curran Associates Inc.
\newblock ISBN 9781510860964.

\bibitem[{Ho, Jain, and Abbeel(2020)}]{ho2020denoising}
Ho, J.; Jain, A.; and Abbeel, P. 2020.
\newblock Denoising diffusion probabilistic models.
\newblock 33: 6840--6851.

\bibitem[{Ho and Salimans(2021)}]{ho2021classifierfree}
Ho, J.; and Salimans, T. 2021.
\newblock Classifier-Free Diffusion Guidance.
\newblock In \emph{NeurIPS 2021 Workshop on Deep Generative Models and Downstream Applications}.

\bibitem[{Hong et~al.(2022)Hong, Zhang, Shen, and Xu}]{hong2022depth}
Hong, F.-T.; Zhang, L.; Shen, L.; and Xu, D. 2022.
\newblock Depth-Aware Generative Adversarial Network for Talking Head Video Generation.

\bibitem[{Isola et~al.(2017)Isola, Zhu, Zhou, and Efros}]{pix2pix2017}
Isola, P.; Zhu, J.-Y.; Zhou, T.; and Efros, A.~A. 2017.
\newblock Image-to-Image Translation with Conditional Adversarial Networks.
\newblock \emph{CVPR}.

\bibitem[{Ji et~al.(2022)Ji, Zhou, Wang, Wu, Wu, Xu, and Cao}]{eamm}
Ji, X.; Zhou, H.; Wang, K.; Wu, Q.; Wu, W.; Xu, F.; and Cao, X. 2022.
\newblock EAMM: One-Shot Emotional Talking Face via Audio-Based Emotion-Aware Motion Model.
\newblock In \emph{ACM SIGGRAPH 2022 Conference Proceedings}, SIGGRAPH '22.

\bibitem[{Karras et~al.(2020)Karras, Laine, Aittala, Hellsten, Lehtinen, and Aila}]{karras2020analyzing}
Karras, T.; Laine, S.; Aittala, M.; Hellsten, J.; Lehtinen, J.; and Aila, T. 2020.
\newblock Analyzing and improving the image quality of stylegan.
\newblock In \emph{Proc. of the {IEEE} {C}onf. on {C}omputer {V}ision and {P}attern {R}ecognition (CVPR)}, 8110--8119.

\bibitem[{Kong et~al.(2021)Kong, Ping, Huang, Zhao, and Catanzaro}]{kong2021diffwave}
Kong, Z.; Ping, W.; Huang, J.; Zhao, K.; and Catanzaro, B. 2021.
\newblock DiffWave: A Versatile Diffusion Model for Audio Synthesis.
\newblock In \emph{{I}nternational {C}onference on {L}earning {R}epresentations (ICLR)}.

\bibitem[{Liu et~al.(2023)Liu, Lin, Fei, Changyin, and Yu}]{liu2023MODA}
Liu, Y.; Lin, L.; Fei, Y.; Changyin, Z.; and Yu, L. 2023.
\newblock MODA: Mapping-Once Audio-driven Portrait Animation with Dual Attentions.
\newblock In \emph{Proceedings of the IEEE/CVF International Conference on Computer Vision}.

\bibitem[{Ma et~al.(2022)Ma, Wang, Petridis, Shen, and Pantic}]{lipreading}
Ma, P.; Wang, Y.; Petridis, S.; Shen, J.; and Pantic, M. 2022.
\newblock Training Strategies for Improved Lip-Reading.
\newblock In \emph{ICASSP 2022 - 2022 IEEE International Conference on Acoustics, Speech and Signal Processing (ICASSP)}, 8472--8476.

\bibitem[{Ma et~al.(2023{\natexlab{a}})Ma, Wang, Hu, Fan, Lv, Ding, Deng, and Yu}]{styletalk}
Ma, Y.; Wang, S.; Hu, Z.; Fan, C.; Lv, T.; Ding, Y.; Deng, Z.; and Yu, X. 2023{\natexlab{a}}.
\newblock StyleTalk: One-Shot Talking Head Generation with Controllable Speaking Styles.
\newblock AAAI'23/IAAI'23/EAAI'23. AAAI Press.

\bibitem[{Ma et~al.(2023{\natexlab{b}})Ma, Zhang, Wang, Wang, Zhang, and Deng}]{ma2023dreamtalk}
Ma, Y.; Zhang, S.; Wang, J.; Wang, X.; Zhang, Y.; and Deng, Z. 2023{\natexlab{b}}.
\newblock DreamTalk: When Expressive Talking Head Generation Meets Diffusion Probabilistic Models.
\newblock \emph{arXiv preprint arXiv:2312.09767}.

\bibitem[{Mao et~al.(2017)Mao, Li, Xie, Lau, Wang, and Paul~Smolley}]{mao2017least}
Mao, X.; Li, Q.; Xie, H.; Lau, R.~Y.; Wang, Z.; and Paul~Smolley, S. 2017.
\newblock Least squares generative adversarial networks.
\newblock In \emph{{P}roceedings of the {IEEE} {I}nternational {C}onference on {C}omputer {V}ision (ICCV)}, 2794--2802.

\bibitem[{Nagrani, Chung, and Zisserman(2017)}]{nagrani2017voxceleb}
Nagrani, A.; Chung, J.; and Zisserman, A. 2017.
\newblock VoxCeleb: a large-scale speaker identification dataset.
\newblock \emph{Interspeech}.

\bibitem[{Pang et~al.(2023)Pang, Zhang, Quan, Fan, Cun, Shan, and Yan}]{Pang_2023_CVPR}
Pang, Y.; Zhang, Y.; Quan, W.; Fan, Y.; Cun, X.; Shan, Y.; and Yan, D.-M. 2023.
\newblock DPE: Disentanglement of Pose and Expression for General Video Portrait Editing.
\newblock In \emph{Proc. of the {IEEE} {C}onf. on {C}omputer {V}ision and {P}attern {R}ecognition (CVPR)}, 427--436.

\bibitem[{Prajwal et~al.(2020)Prajwal, Mukhopadhyay, Namboodiri, and Jawahar}]{wav2lip}
Prajwal, K.~R.; Mukhopadhyay, R.; Namboodiri, V.~P.; and Jawahar, C. 2020.
\newblock A Lip Sync Expert Is All You Need for Speech to Lip Generation In the Wild.
\newblock MM '20, 484–492. Association for Computing Machinery.

\bibitem[{Ramesh et~al.(2022)Ramesh, Dhariwal, Nichol, Chu, and Chen}]{ramesh2022hierarchical}
Ramesh, A.; Dhariwal, P.; Nichol, A.; Chu, C.; and Chen, M. 2022.
\newblock Hierarchical Text-Conditional Image Generation with CLIP Latents.
\newblock \emph{arXiv e-prints}, arXiv--2204.

\bibitem[{Ren et~al.(2021)Ren, Li, Chen, Li, and Liu}]{Ren_2021_ICCV}
Ren, Y.; Li, G.; Chen, Y.; Li, T.~H.; and Liu, S. 2021.
\newblock PIRenderer: Controllable Portrait Image Generation via Semantic Neural Rendering.
\newblock In \emph{{P}roceedings of the {IEEE} {I}nternational {C}onference on {C}omputer {V}ision (ICCV)}, 13759--13768.

\bibitem[{Saharia et~al.(2022)Saharia, Chan, Saxena, Li, Whang, Denton, Ghasemipour, Gontijo~Lopes, Karagol~Ayan, Salimans et~al.}]{saharia2022photorealistic}
Saharia, C.; Chan, W.; Saxena, S.; Li, L.; Whang, J.; Denton, E.~L.; Ghasemipour, K.; Gontijo~Lopes, R.; Karagol~Ayan, B.; Salimans, T.; et~al. 2022.
\newblock Photorealistic text-to-image diffusion models with deep language understanding.
\newblock \emph{Advances in Neural Information Processing Systems}, 35: 36479--36494.

\bibitem[{Shen et~al.(2023)Shen, Zhao, Meng, Li, Zhu, Zhou, and Lu}]{shen2023difftalk}
Shen, S.; Zhao, W.; Meng, Z.; Li, W.; Zhu, Z.; Zhou, J.; and Lu, J. 2023.
\newblock DiffTalk: Crafting Diffusion Models for Generalized Audio-Driven Portraits Animation.
\newblock In \emph{Proc. of the {IEEE} {C}onf. on {C}omputer {V}ision and {P}attern {R}ecognition (CVPR)}.

\bibitem[{Siarohin et~al.(2019)Siarohin, Lathuilière, Tulyakov, Ricci, and Sebe}]{Siarohin_2019_NeurIPS}
Siarohin, A.; Lathuilière, S.; Tulyakov, S.; Ricci, E.; and Sebe, N. 2019.
\newblock First Order Motion Model for Image Animation.

\bibitem[{Siarohin et~al.(2021)Siarohin, Woodford, Ren, Chai, and Tulyakov}]{siarohin2021motion}
Siarohin, A.; Woodford, O.; Ren, J.; Chai, M.; and Tulyakov, S. 2021.
\newblock Motion Representations for Articulated Animation.
\newblock In \emph{Proc. of the {IEEE} {C}onf. on {C}omputer {V}ision and {P}attern {R}ecognition (CVPR)}.

\bibitem[{Simonyan and Zisserman(2015)}]{Simonyan15}
Simonyan, K.; and Zisserman, A. 2015.
\newblock Very Deep Convolutional Networks for Large-Scale Image Recognition.
\newblock In \emph{International Conference on Learning Representations}.

\bibitem[{Song et~al.(2019)Song, Zhu, Li, Wang, and Qi}]{ijcai2019p129}
Song, Y.; Zhu, J.; Li, D.; Wang, A.; and Qi, H. 2019.
\newblock Talking Face Generation by Conditional Recurrent Adversarial Network.
\newblock 919--925. International Joint Conferences on Artificial Intelligence Organization.

\bibitem[{Tian et~al.(2024)Tian, Wang, Zhang, and Bo}]{tian2024emo}
Tian, L.; Wang, Q.; Zhang, B.; and Bo, L. 2024.
\newblock EMO: Emote Portrait Alive -- Generating Expressive Portrait Videos with Audio2Video Diffusion Model under Weak Conditions.
\newblock arXiv:2402.17485.

\bibitem[{Vougioukas, Petridis, and Pantic(2019)}]{Vougioukas2019EndtoEndSR}
Vougioukas, K.; Petridis, S.; and Pantic, M. 2019.
\newblock End-to-End Speech-Driven Realistic Facial Animation with Temporal GANs.
\newblock In \emph{Proc. of the {IEEE} {I}nternational {C}onference on {C}omputer {V}ision {W}orkshops}.

\bibitem[{Wang et~al.(2021{\natexlab{a}})Wang, Li, Ding, Fan, and Yu}]{wang2021audio2head}
Wang, S.; Li, L.; Ding, Y.; Fan, C.; and Yu, X. 2021{\natexlab{a}}.
\newblock Audio2Head: Audio-driven One-shot Talking-head Generation with Natural Head Motion.

\bibitem[{Wang et~al.(2022{\natexlab{a}})Wang, Li, Ding, and Yu}]{wang2021one}
Wang, S.; Li, L.; Ding, Y.; and Yu, X. 2022{\natexlab{a}}.
\newblock One-shot Talking Face Generation from Single-speaker Audio-Visual Correlation Learning.

\bibitem[{Wang, Mallya, and Liu(2021)}]{wang2021facevid2vid}
Wang, T.-C.; Mallya, A.; and Liu, M.-Y. 2021.
\newblock One-Shot Free-View Neural Talking-Head Synthesis for Video Conferencing.
\newblock In \emph{Proc. of the {IEEE} {C}onf. on {C}omputer {V}ision and {P}attern {R}ecognition (CVPR)}.

\bibitem[{Wang et~al.(2021{\natexlab{b}})Wang, Li, Zhang, and Shan}]{wang2021gfpgan}
Wang, X.; Li, Y.; Zhang, H.; and Shan, Y. 2021{\natexlab{b}}.
\newblock Towards Real-World Blind Face Restoration with Generative Facial Prior.
\newblock In \emph{The IEEE Conference on Computer Vision and Pattern Recognition (CVPR)}.

\bibitem[{Wang et~al.(2022{\natexlab{b}})Wang, Yang, Bremond, and Dantcheva}]{wang2022latent}
Wang, Y.; Yang, D.; Bremond, F.; and Dantcheva, A. 2022{\natexlab{b}}.
\newblock Latent Image Animator: Learning to Animate Images via Latent Space Navigation.
\newblock In \emph{{I}nternational {C}onference on {L}earning {R}epresentations (ICLR)}.

\bibitem[{Wang et~al.(2004)Wang, Bovik, Sheikh, and Simoncelli}]{ssim}
Wang, Z.; Bovik, A.; Sheikh, H.; and Simoncelli, E. 2004.
\newblock Image quality assessment: from error visibility to structural similarity.
\newblock \emph{{IEEE} {T}rans. on {I}mage {P}rocessing (TIP)}, 13(4): 600--612.

\bibitem[{Xu et~al.(2024)Xu, Chen, Guo, Yang, Li, Zang, Zhang, Tong, and Guo}]{xu2024vasa1}
Xu, S.; Chen, G.; Guo, Y.-X.; Yang, J.; Li, C.; Zang, Z.; Zhang, Y.; Tong, X.; and Guo, B. 2024.
\newblock VASA-1: Lifelike Audio-Driven Talking Faces Generated in Real Time.
\newblock arXiv:2404.10667.

\bibitem[{Yin et~al.(2022)Yin, Zhang, Cun, Cao, Fan, Wang, Bai, Wu, Wang, and Yang}]{yin2022styleheat}
Yin, F.; Zhang, Y.; Cun, X.; Cao, M.; Fan, Y.; Wang, X.; Bai, Q.; Wu, B.; Wang, J.; and Yang, Y. 2022.
\newblock Styleheat: One-shot high-resolution editable talking face generation via pre-trained stylegan.
\newblock In \emph{{P}roceedings of the {E}uropean {C}onference on {C}omputer {V}ision (ECCV)}, 85--101. Springer.

\bibitem[{Yu et~al.(2018)Yu, Wang, Peng, Gao, Yu, and Sang}]{yu2018bisenet}
Yu, C.; Wang, J.; Peng, C.; Gao, C.; Yu, G.; and Sang, N. 2018.
\newblock Bisenet: Bilateral segmentation network for real-time semantic segmentation.
\newblock In \emph{Proceedings of the European conference on computer vision (ECCV)}, 325--341.

\bibitem[{Yu et~al.(2023)Yu, Yin, Zhou, Wang, Wong, and Wang}]{yu2023thpad}
Yu, Z.; Yin, Z.; Zhou, D.; Wang, D.; Wong, F.; and Wang, B. 2023.
\newblock Talking Head Generation with Probabilistic Audio-to-Visual Diffusion Priors.
\newblock In \emph{{P}roceedings of the {IEEE} {I}nternational {C}onference on {C}omputer {V}ision (ICCV)}.

\bibitem[{Zhang et~al.(2023)Zhang, Cun, Wang, Zhang, Shen, Guo, Shan, and Wang}]{Zhang_2023_CVPR}
Zhang, W.; Cun, X.; Wang, X.; Zhang, Y.; Shen, X.; Guo, Y.; Shan, Y.; and Wang, F. 2023.
\newblock SadTalker: Learning Realistic 3D Motion Coefficients for Stylized Audio-Driven Single Image Talking Face Animation.
\newblock In \emph{Proc. of the {IEEE} {C}onf. on {C}omputer {V}ision and {P}attern {R}ecognition (CVPR)}, 8652--8661.

\bibitem[{Zhang et~al.(2021)Zhang, Li, Ding, and Fan}]{zhang2021flow}
Zhang, Z.; Li, L.; Ding, Y.; and Fan, C. 2021.
\newblock Flow-Guided One-Shot Talking Face Generation With a High-Resolution Audio-Visual Dataset.
\newblock In \emph{Proc. of the {IEEE} {C}onf. on {C}omputer {V}ision and {P}attern {R}ecognition (CVPR)}, 3661--3670.

\bibitem[{Zhao and Zhang(2022)}]{zhao2022thin}
Zhao, J.; and Zhang, H. 2022.
\newblock Thin-plate spline motion model for image animation.
\newblock In \emph{Proc. of the {IEEE} {C}onf. on {C}omputer {V}ision and {P}attern {R}ecognition (CVPR)}, 3657--3666.

\bibitem[{Zhou et~al.(2019)Zhou, Liu, Liu, Luo, and Wang}]{zhou2018talking}
Zhou, H.; Liu, Y.; Liu, Z.; Luo, P.; and Wang, X. 2019.
\newblock Talking Face Generation by Adversarially Disentangled Audio-Visual Representation.

\bibitem[{Zhou et~al.(2021)Zhou, Sun, Wu, Loy, Wang, and Liu}]{zhou2021pose}
Zhou, H.; Sun, Y.; Wu, W.; Loy, C.~C.; Wang, X.; and Liu, Z. 2021.
\newblock Pose-Controllable Talking Face Generation by Implicitly Modularized Audio-Visual Representation.
\newblock In \emph{Proc. of the {IEEE} {C}onf. on {C}omputer {V}ision and {P}attern {R}ecognition (CVPR)}.

\bibitem[{Zhou et~al.(2020)Zhou, Han, Shechtman, Echevarria, Kalogerakis, and Li}]{zhou2020makelttalk}
Zhou, Y.; Han, X.; Shechtman, E.; Echevarria, J.; Kalogerakis, E.; and Li, D. 2020.
\newblock Makelttalk: speaker-aware talking-head animation.
\newblock \emph{{ACM} {T}rans. on {G}raphics (TOG)}, 39(6): 1--15.

\end{thebibliography}

\clearpage
\appendix

\begin{appendices}
\section*{Supplementary Materials}
In this exposition, we present more implementation details and experimental results to support our work. Concretely, the following content comprising of
\begin{enumerate}
\item implementation details with respect to the preliminary of intermediate 3D motion descriptors, loss functions of each proposed module and corresponding ablation studies,
\item supplementary experiments on several highlighted aspects in the main paper,
\item discussion about limitations and future directions, along with ethical considerations
\end{enumerate} are reported.

\section{Preliminary of 3D Motion Descriptors}
Explicit representations for modeling facial motions, e.g., facial landmarks and 3DMM, are advantageous for intuitive motion editing. Notably, the 3DMM coefficients naturally decouple head poses and facial expressions, which is beneficial for separately modeling pose and expression generation, as they have distinct correlations with the audio. Therefore, we employ a subset of 3DMM coefficients as the intermediate motion descriptors for facial animation, with which the shape $\mathbf{S}$ of a face can be parameterized as:
\begin{equation}
    \label{eq:3dmm} \mathbf{S}=\bar{\mathbf{S}} + \mathbf{i} \mathbf{B}_{id} + \mathbf{e} \mathbf{B}_{exp},
\end{equation}
where $\bar{\mathbf{S}}$ is the average face shape, $\mathbf{B}_{id}$ and $\mathbf{B}_{exp}$ are the basis of identity and expression. The coefficients $\mathbf{i} \in \mathbb{R}^{80}$ and $\mathbf{e} \in \mathbb{R}^{64}$ represents the facial shape and expression, respectively. Moreover, the head poses are described by $\mathbf{p}=\mathbf{r}\oplus\mathbf{t}$, where $\oplus$ means concatenation (through out the paper), $\mathbf{r} \in SO(3)$ is the rotation vector and $\mathbf{t} \in \mathbb{R}^3$ is the transformation vector. We adopt $\mathbf{p}$ and $\mathbf{e}$ as the intermediate representations for facial motion modeling and generation. As for the identity information, we directly use the input source image for more precise identity and texture reference in the final facial animation process. In addition, we incorporate two values to represent the pitch and yaw angles of gazes, which is denoted as $\mathbf{g} \in \mathbb{R}^{2}$. 

\section{Loss Functions and Ablation Study}
\subsection{Loss Functions for the Face Animator}
The training objective of our face animator consists of the following parts:

\noi{Reconstruction Loss $\mathcal{L}_{rec}$.} Basically, the mean absolute errors between the generated and ground-truth target images are calculated. To ensure precise transformation of mouth and eye motions, facial landmark detection is utilized to derive bounding boxes for the corresponding regions. Extra weights are assigned during pixel-wise loss computation within these delineated areas, thereby intensifying the focus on these specific regions in the overall loss calculation process:
\begin{equation}
    \mathcal{L}_{rec}(I_t^D, \hat{I}_t^D)=\mathbb{E}[\Vert I_t^D-\hat{I}_t^D\Vert_1 \odot (1+M_{lip\&eye})],
\end{equation}
where $M_{lip\&eye}$ is the weighted mask emphasizing the lip and eye regions, and $\odot$ means element-wise multiplication (throughout the paper).

\noi{Perceptual Loss $\mathcal{L}_{perc}$.} To increase the realism of the animated images, we employ a masked multi-scale perceptual loss based on a pre-trained VGG19 network \cite{Simonyan15}:

\begin{equation}
    \resizebox{0.55\textwidth}{!}{$\mathcal{L}_{perc}(I_t^D, \hat{I}_t^D)=\mathbb{E}[\sum_n^N\Vert \Phi_{VGG}^n(I_t^D)-\Phi_{VGG}^n(\hat{I}_t^D)\Vert_1 \odot (1+M_{lip\&eye})],$}
\end{equation}
where $\Phi_{VGG}^n$ denotes the \emph{n}-th layer of the VGG network. Practically, we leverage the feature maps of four resolutions: $256\times 256$, $128\times 128$, $64\times 64$, and $32\times 32$.

\noi{Expression Loss $\mathcal{L}_{exp}$.} To ensure accurate control on expressions, a pre-trained expression recognition network \cite{DECA} is applied to calculate the distances between expression features as expression loss:
\begin{equation}
    \mathcal{L}_{exp}(I_t^D, \hat{I}_t^D)=\mathbb{E}[\Vert \Phi_{exp}(I_t^D)-\Phi_{exp}(\hat{I}_t^D)\Vert_1],
\end{equation}
where $\Phi_{exp}$ indicates the feature extracting layers of the expression recognition network.

\noi{Gaze Loss $\mathcal{L}_{gaze}$.} Towards precise animation on gazes, we use a pre-trained gaze estimator \cite{Abdelrahman2022L2CSNetFG} to extract gaze features and then compute the gaze loss:
\begin{equation}
    \mathcal{L}_{gaze}(I_t^D, \hat{I}_t^D)=\mathbb{E}[\Vert \Phi_{gaze}(I_t^D)-\Phi_{gaze}(\hat{I}_t^D)\Vert_1],
\end{equation}
where $\Phi_{gaze}$ is the encoding layers of the gaze estimator.

\begin{figure}[t]
	\begin{subfigure}{1.\linewidth}
        \centering
			\begin{minipage}{0.4\linewidth}
                \centering
			    \includegraphics[width=0.9\linewidth]{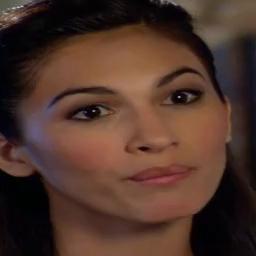}  
                \caption{Input}
			\end{minipage}
			\begin{minipage}{0.4\linewidth}
                \centering
			    \includegraphics[width=0.9\linewidth]{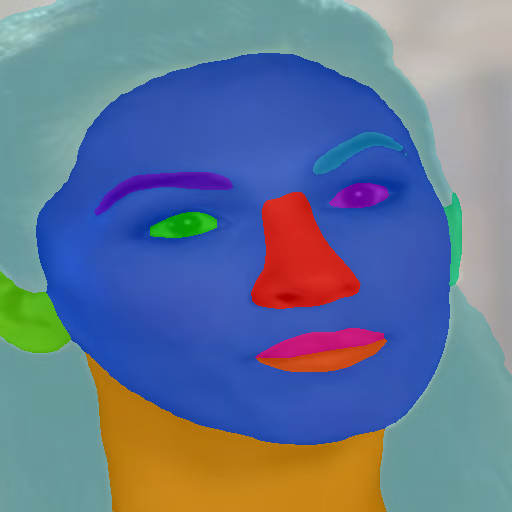}
                \caption{Parsed}
			\end{minipage}
	\end{subfigure}
    \caption{Illustration of semantic segmentation with a pre-trained BiSeNet \cite{yu2018bisenet} model.}
    \label{fig:parsing}
\end{figure}

\noi{Parsing Loss $\mathcal{L}_{pars}$.} To increase the semantic sensitivity of our face animator, we also exert a semantic segmentation model BiSeNet \cite{yu2018bisenet} on the generated results to help the network distinguish different facial regions and additionally background and foreground:
\begin{equation}
    \mathcal{L}_{pars}(I_t^D, \hat{I}_t^D)=PCE(\Phi_{pars}(I_t^D), \Phi_{pars}(\hat{I}_t^D)),
\end{equation}
where $\Phi_{pars}$ denotes the pre-trained parsing network utilized for semantic segmentation, and $PCE$ corresponds to the operation of computing pixel-wise cross-entropy loss. An example of the parsing result is shown in Fig. \ref{fig:parsing}. 

\noi{Adversarial Loss $\mathcal{L}_{GAN}$.} Finally, to generate photo-realistic results, we adopt the non-saturating adversarial loss as our adversarial loss:
\begin{equation}
    \mathcal{L}_{GAN}(\hat{I}_t^D)=\mathbb{E}[-\log D(\hat{I}_t^D))],
\end{equation}
where $D$ is a discriminator for distinguishing reconstructed images from the real ones.

In summary, the total loss of our face animator can be calculated as:
\begin{align}
    \notag \mathcal{L}_{total}=\lambda_{rec}\mathcal{L}_{rec}+\lambda_{perc}\mathcal{L}_{perc}+\lambda_{exp}\mathcal{L}_{exp}\\
    \label{eq:face_ani_loss}+\lambda_{gaze}\mathcal{L}_{gaze}+\lambda_{pars}\mathcal{L}_{pars}+\mathcal{L}_{GAN}.
\end{align}
In our implementation, the $\lambda$s are assigned by: $\lambda_{rec}=1$, $\lambda_{perc}=1$, $\lambda_{exp}=1$, $\lambda_{gaze}=100$, and $\lambda_{pars}=1$. 
\\\\
\noi{Ablation Study.} 
We conduct ablation studies on the parsing loss $\mathcal{L}_{pars}$ and the employed partial mask $M_{lip\&eye}$ to validate their significance, as they may not be considered necessities. Evaluations of lip synchronization and the quality of the final generated videos through training strategies without one of the mentioned modules (w/o $\mathcal{L}_{pars}$ and w/o $M_{lip\&eye}$) are reported in Table \ref{tab:face_ab}. With attention focused on the lip region through parsing and specific masks, the \textbf{Full} strategy outperforms in lip synchronization with almost negligible loss in video qualities. Examples in our supplementary video also demonstrate its superiority in learning the overall facial fidelity and distinguishing the foreground and the background.
\begin{table}[htbp]
  \centering
   \resizebox{\linewidth}{!}{
    \renewcommand\arraystretch{1.1}
    \begin{tabular}{l|ccc|ccc}
    \toprule
    \multirow{2}[4]{*}{Strategy} & \multicolumn{3}{c|}{HDTF} & \multicolumn{3}{c}{VoxCeleb} \\
\cmidrule{2-7}          & LSE-C $\uparrow$& LSE-D $\downarrow$ & FID $\downarrow$   & LSE-C $\uparrow$ & LSE-D $\downarrow$ & FID $\downarrow$\\
    \cline{1-7}
    \cline{1-7}
    \cline{1-7}
    w/o $\mathcal{L}_{pars}$ & 7.96  & 8.10  & \textbf{27.32} & 7.13  & 8.30  & 58.23 \\
    w/o $M_{lip\&eye}$ & 7.90  & \textbf{7.77}  & 28.56 & 6.98  & 8.23  & 58.93 \\
    \textbf{Full}  & \textbf{8.13}  & 7.78  & 30.40 & \textbf{7.20}  & \textbf{8.19}  & \textbf{58.11} \\
    \bottomrule
    \end{tabular}}%
    \caption{Ablation study for loss functions of the face animator.}
  \label{tab:face_ab}%
\end{table}%

\begin{figure}
    \centering
    \includegraphics[width=.5\linewidth]{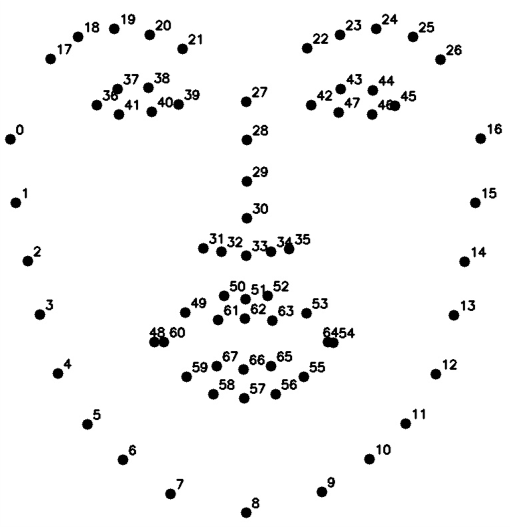}
    \caption{Landmarks indices visualization.}
    \label{fig:lms}
\end{figure}

\subsection{Loss Functions and Training Strategies for the Expression Predictor}
Taking the videos resynchrinzed through a pre-trained lip expert \cite{wav2lip} as learning targets, the first stage utilizes supervised training with specially designed loss functions to facilitate the distillation of precise audio-to-lip mapping and eye motion maintenance. Specifically, we apply L1 losses on the coefficients ($\mathcal{L}_1$) and their corresponding eye motion features ($\mathcal{L}_{eye}$). Additionally, to ensure more accurate opening and closing of the lip shape, we introduce the landmarks fidelity loss $\mathcal{L}_{lms}$ around the lip and jaw area and a "shut up" loss $\mathcal{L}_{shut}$ based on the diagonal distances between the upper and lower inner lip landmarks. Moreover, a lip reading loss $\mathcal{L}_{read}$ is incorporated by leveraging a lip reading expert $\mathbf{\Phi}_{read}$ \cite{lipreading} on the rendered images of facial meshes reconstructed from the expression coefficients. In summary, the loss function for the first-stage training can be written as:
\begin{align}
    \label{eq:stage1}\mathcal{L}_{stage1}&=\mathcal{L}_{distill}, \\
    \notag \mathcal{L}_{distill}&=\lambda_{lms}\mathcal{L}_{lms}+\lambda_{eye}\mathcal{L}_{eye}+\lambda_{shut}\mathcal{L}_{shut}\\
    \label{eq:distill}&+\lambda_{read}\mathcal{L}_{read}+\lambda_1\mathcal{L}_1,\\
    \label{eq:lip}\mathcal{L}_{lms}&=\frac{1}{T}\sum_{t=1}^{T}\frac{1}{31}\sum_{\substack{i\in[3,14]\cup[48,67]}}\left\Vert \boldsymbol{l}^i_t-\tilde{\boldsymbol{l}}^i_t \right\Vert^2_2, \\
    \label{eq:eye}\mathcal{L}_{eye} &= \frac{1}{T}\sum_{t=1}^{T}\left\Vert \boldsymbol{f}_t - \tilde{\boldsymbol{f}}_t\right\Vert_1,\\
    \notag\mathcal{L}_{shut}&=\frac{1}{T}\sum_{t=1}^{T}\left\Vert \boldsymbol{d}_t-\tilde{\boldsymbol{d}}_t \right\Vert^2_2\\
    \label{eq:shut}&=\left\Vert \left(\boldsymbol{l}^{60:63}_t-\boldsymbol{l}^{64:67}_t\right)-\left(\tilde{\boldsymbol{l}}^{60:63}_t-\tilde{\boldsymbol{l}}^{64:67}_t\right) \right\Vert^2_2,\\
    \label{eq:read}\mathcal{L}_{read}&=\frac{1}{T}\sum^T_{t=1}\left(1-\frac{\mathbf{\Phi}_{read}(\boldsymbol{r}_t)\cdot \mathbf{\Phi}_{read}(\tilde{\boldsymbol{r}}_{t})}{\max{(\Vert\mathbf{\Phi}_{read}(\boldsymbol{r}_t)\Vert_2 \cdot \Vert\mathbf{\Phi}_{read}(\tilde{\boldsymbol{r}}_{t})\Vert_2, \epsilon)}}\right) ,\\
    \label{eq:expl1}\mathcal{L}_1&=\frac{1}{T}\sum_{t=1}^{T}\left\Vert\mathbf{e}_t - \tilde{\mathbf{e}}_t\right\Vert_1,
\end{align}
where the characters with and without the tilde notation $\tilde{}$ respectively represent the corresponding predicted and ground-truth values. $\boldsymbol{l}^i_t$ denotes the \emph{i}-th landmark at the \emph{t}-th frame, and $\boldsymbol{r}_t$ is the cropped mouth area of the image rendered from $\mathbf{e}_t$. Unlike the 2D landmarks used for extracting vertical eye motion features, we apply 3D landmarks for loss calculation to encourage the network to learn lip motions from more spatial dimensions.

With the learned weights of $\boldsymbol{E}_{map}$ frozen, in the second stage, we employ adversarial training for the LSTM-based generator. In addition to the well-designed loss function $\mathcal{L}_{stage1}$ for mapping lip motions and learning coefficients fidelity, we introduce three discriminators, including an eye motion discriminator ($\boldsymbol{D}_{eye}$), and two temporal discriminators ($\boldsymbol{D}_{te}$ and $\boldsymbol{D}_{tf}$). In this stage, we use the hat notation $\hat{}$ to represent generated values. Moreover, Structural Similarity \cite{ssim} loss ($\mathcal{L}_{ssim}$) and L1 loss ($\mathcal{L}'_1$) are also implemented to help the network understand the statistical structures of eye motions and generate results close to the ground-truth distribution. Therefore, the loss function for the second stage can be summarized as follows:
\begin{align}
    \notag \mathcal{L}_{stage2}&=\lambda_{ssim}\mathcal{L}_{ssim}+\lambda'_1\mathcal{L}'_1+\lambda_{ad}\mathcal{L}_{ad}(\boldsymbol{G}_{lstm}, \boldsymbol{D}_{eye})\\
    \notag &+\lambda_{te}\mathcal{L}_{te}(\boldsymbol{G}_{lstm}, \boldsymbol{D}_{te})+\lambda_{tf}\mathcal{L}_{tf}(\boldsymbol{G}_{lstm}, \boldsymbol{D}_{tf})\\
    \label{eq:stage2} &+\lambda_{stage1}\mathcal{L}_{stage1}, \\
    \label{eq:ssim}\mathcal{L}_{ssim}&=1-\frac{1}{21}\sum_{\substack{i\in[0,20]}}\frac{(2\mu_i\hat{\mu}_i+C_1)(2cov_i+C_2)}{(\mu_i^2+\hat{\mu}_i^2+C_1)(\sigma_i^2+\hat{\sigma}_i^2+C_2)}, \\
    \label{eq:eyel1} \mathcal{L}'_1 &= \frac{1}{T}\sum_{t=1}^{T}\Vert \boldsymbol{f}_t - \hat{\boldsymbol{f}}_t\Vert_1,
\end{align}
where $\boldsymbol{D}_{eye}$ uses a transformer-based structure with both the sequences of audio features and landmarks around the eyes area ($\boldsymbol{l}^{17:26}_{1:T}\oplus\boldsymbol{l}^{36:47}_{1:T}/\hat{\boldsymbol{l}}^{17:26}_{1:T}\oplus\hat{\boldsymbol{l}}^{36:47}_{1:T}$) as inputs; $\boldsymbol{D}_{te}$ and $\boldsymbol{D}_{tf}$ follows the structure of PatchGAN \cite{pix2pix2017}. The three discriminators are all trained jointly with the generator using least square losses \cite{mao2017least}. $C_1$ and $C_2$ are two small constants, $(\hat{\mu}_i,\mu_i)$ and $(\hat{\sigma}_i,\sigma_i)$ are the mean and standard deviation of the \emph{i}-th dimension of $(\hat{\boldsymbol{f}}_{1:T},\boldsymbol{f}_{1:T})$, with $cov_i$ being the covariance. The indices of facial landmarks are visualized in Fig. \ref{fig:lms}.
\\\\
\noi{Ablation Study.}
In the first stage, we perform ablation studies on the HDTF dataset to validate the impact of the loss functions on lip synchronization. The mouth landmark distances (MLD), derived from the predicted expression coefficients under different strategies, are presented in Table \ref{tab:stage1_ab}. Results without any of the mouth-related loss terms demonstrate poorer performances than the \textbf{Full} strategy, indicating the crucial role of these loss functions in enhancing lip synchronization accuracy.

\begin{table}[htbp]
  \centering
    \renewcommand\arraystretch{1.1}
    \begin{tabular}{l|c}
    \toprule
    Strategy & MLD $\downarrow$ \\
    \cline{1-2}
    \cline{1-2}
    \cline{1-2}
    w/o $\mathcal{L}_{read}$ & 1.76 \\
    w/o $\mathcal{L}_{lms}$ & 1.79 \\
    w/o $\mathcal{L}_{shut}$ & 1.72 \\
    \textbf{Full} & \textbf{1.65} \\
    \bottomrule
    \end{tabular}%
    \caption{Ablation study for loss functions in the first stage of expression predicted.}
  \label{tab:stage1_ab}%
\end{table}%

In the second stage, we conduct ablation studies on 100 videos of the HDTF dataset with the following variants: 1) w/o $\mathcal{L}_{ssim}$: Remove the structural similarity loss in the second stage. 2)  w/o $D_{eye}$: Exclude the eye motion discriminator. 3)  w/o $D_{tf}$: Exclude the temporal discriminator for coefficient sequences. 4)  w/o $D_{te}$: Exclude the temporal discriminator for eye motion sequences. 5) \textbf{Full}: Our full training strategy.

We compute the average mouth landmarks distances (MLD) and eye motion structural similarities ($\text{SSIM}_e$) for the generated expression coefficients to observe the effect of each module in lip synchronization and eye motion generation. The results are demonstrated in Table \ref{tab:twostage_ab}. The $\text{SSIM}_e$ scores reduce significantly after removing any of the proposed modules, indicating their significance in facilitating data structural fidelity. However, there appears to be a trade-off between generating a realistic eye motion sequence and mapping precise lip motions, as all strategies present higher MLD than that of only training the first stage (strategies in Table \ref{tab:stage1_ab}). Despite the potential loss in lip movements, our \textbf{Full} strategy still excels in maintaining the realism of generated eye motions.

\begin{table}[t]
  \centering
    \renewcommand\arraystretch{1.1}
    \begin{tabular}{l|ccc}
    \toprule
    Strategy & MLD $\downarrow$   & $\text{SSIM}_e$ $\uparrow$\\
    \cline{1-3}
    \cline{1-3}
    \cline{1-3}
    w/o $\mathcal{L}_{ssim}$ &  1.870  &  0.845 \\
    w/o $D_{eye}$ &   \textbf{1.785}   &  0.840   \\
    w/o $D_{tf}$ &   1.832    &   0.844  \\
    w/o $D_{te}$ &   1.846    &   0.842  \\
    \textbf{Full} & 1.792 &  \textbf{0.915} \\
    \bottomrule
    \end{tabular}%
    \caption{Ablation results for the second stage in expression prediction. The best results are highlighted in \textbf{bold}.}
  \label{tab:twostage_ab}%
\end{table}%

\section{Implementation Details}
A subset of the preprocessed VoxCeleb \cite{nagrani2017voxceleb} dataset containing over 100k videos of over 1k identities is used for training our face animator. For synthesizing head poses and variations, we employ a part of the HDTF \cite{zhang2021flow} dataset as the training data to learn speech-stylized head motions and realistic gazes. The estimation of expressions requires highly aligned audio and visual data, thus we utilize the pre-trained lip-synchronization model \cite{wav2lip} to generate audio-aligned videos from the initial HDTF videos. Then we extract audio-expression sequences for training the two-stage audio-to-expression model. The testing data are randomly selected from the untrained parts of the HDTF dataset and the VoxCeleb dataset with 220 clipped videos of 16 identities from the former and 397 videos from the latter. The first frame of each video is utilized as the source input image for talking face animation, while the input audios are extracted from the testing videos and down-sampled to 16kHz. 

The training of all models is conducted on four GeForce RTX 3090 GPUs. Specifically, all models use the Adam optimizer and the initial learning rates are set to $6e^{-4}$, $1e^{-4}$, $1e^{-6}$, and $1e^{-4}$ for the pose denoising network, both stages of the two-stage expression predictor, the discriminators for eye motions generation, and the face animator, respectively. The coefficient $\gamma$s in pose synthesis are set to $0.4$ to find a balance between generation diversity and the influence of input signals ($\mathbf{p}_0/\mathbf{g}_0$). The training sequence lengths are set to 30 and 300 respectively for the first and second stages of expression generation, as the former focuses on frame-wise alignment while the latter requires dynamic reliance on longer time scales. 
The $\lambda$s in Eq.(\ref{eq:stage1}) and Eq.(\ref{eq:stage2}) are assigned as follows: $\lambda_{lms}=0.01$, $\lambda_{eye}=2$, $\lambda_{shut}=1$, $\lambda_{read}=2$, $\lambda_1=2$, $\lambda_{ssim}=2$, $\lambda'_1=2$, $\lambda_{ad}=1$, $\lambda_{te}=1$, $\lambda_{tf}=1$, and $\lambda_{stage1}=1$.

\section{Supplementary Experiments}


\subsection{Conformer vs. Convolution in $\boldsymbol{\epsilon}_\theta$} 
We conduct ablation studies on the HDTF and the VoxCeleb dataset to validate the rationale of incorporating attention modules into the denoising network $\boldsymbol{\epsilon}_\theta$ in pose and gaze synthesis. Using the generated poses as representatives, we calculate their average structural similarities to the ground truth poses, as illustrated in Table \ref{tab:att_ab}. After removing the attention modules (w/o Attention), the scores decrease remarkably, underscoring the importance of attention in learning data structural fidelity and generating more realistic results. Users can refer to the supplementary video for a sight of more intuitive demonstration.
\begin{table}[t]
  \centering  
    \renewcommand\arraystretch{1.1}
    \begin{tabular}{l|c|c}
    \toprule
    \multirow{2}[4]{*}{Strategy} & \multicolumn{2}{c}{$\text{SSIM}_p$ $\uparrow$} \\
\cmidrule{2-3}          & HDTF  & VoxCeleb \\
    \cline{1-3}
    \cline{1-3}
    \cline{1-3}
    w/o Attention &  0.983  &  0.933 \\
    \textbf{Full}  &  \textbf{0.996}  &  \textbf{0.987} \\
    \bottomrule
    \end{tabular}%
    \caption{Ablation study for whether incorporating attention modules into the convolution-based denoising network $\boldsymbol{\epsilon}_\theta$.}
  \label{tab:att_ab}%
\end{table}%

\begin{table}[htbp]
  \centering
    \renewcommand\arraystretch{1.1}
    \begin{tabular}{c|c|c}
    \toprule
    \multirow{2}[4]{*}{Method} & \multicolumn{2}{c}{CSIM $\uparrow$} \\
\cmidrule{2-3}          & HDTF  & VoxCeleb \\
    \cline{1-3}
    \cline{1-3}
    \cline{1-3}
    MakeItTalk \shortcite{zhou2020makelttalk} & 0.747 & 0.667 \\
    Wav2Lip \shortcite{wav2lip} & 0.762 & 0.676 \\
    Audio2Head \shortcite{wang2021audio2head} & 0.692 & 0.424 \\
    EAMM \shortcite{eamm} & 0.677 & 0.580 \\
    SadTalker \shortcite{Zhang_2023_CVPR} & 0.766 & \textbf{0.763} \\
    Ours  & 0.765 & 0.684 \\
    \cline{1-3}
    Ground Truth    & \textbf{0.808} & 0.696 \\
    \bottomrule
    \end{tabular}%
    \caption{Assessment of identity preservation, where we compare the cosine similarities (CSIM) of identity features extracted by ArcFace \cite{deng2019arcface} on both the HDTF dataset and the VoxCeleb dataset.}
  \label{tab:csim}%
\end{table}%

\subsection{Identity Preservation}
We also compare the identity-preserving abilities of different methods on both datasets. Using an off-the-shelf face recognition model \cite{deng2019arcface}, we calculate the cosine similarities (CSIM) of identity embeddings between the source images and the generated frames. The results are presented in Table \ref{tab:csim}. It is important to note that pose and expressions also influence identification, as the model relies partly on the landmarks of the input image. Therefore, a higher score may result not only from a similar identity but also from small variations in head and facial motions. According to this assumption, our method achieves a score relatively similar to the ground truth in both datasets, indicating a notable preservation of identity despite potential variations in head and facial motions.

\subsection{Out-of-Distribution Animation}
\begin{figure}[t]
    \centering
    \includegraphics[width=\linewidth]{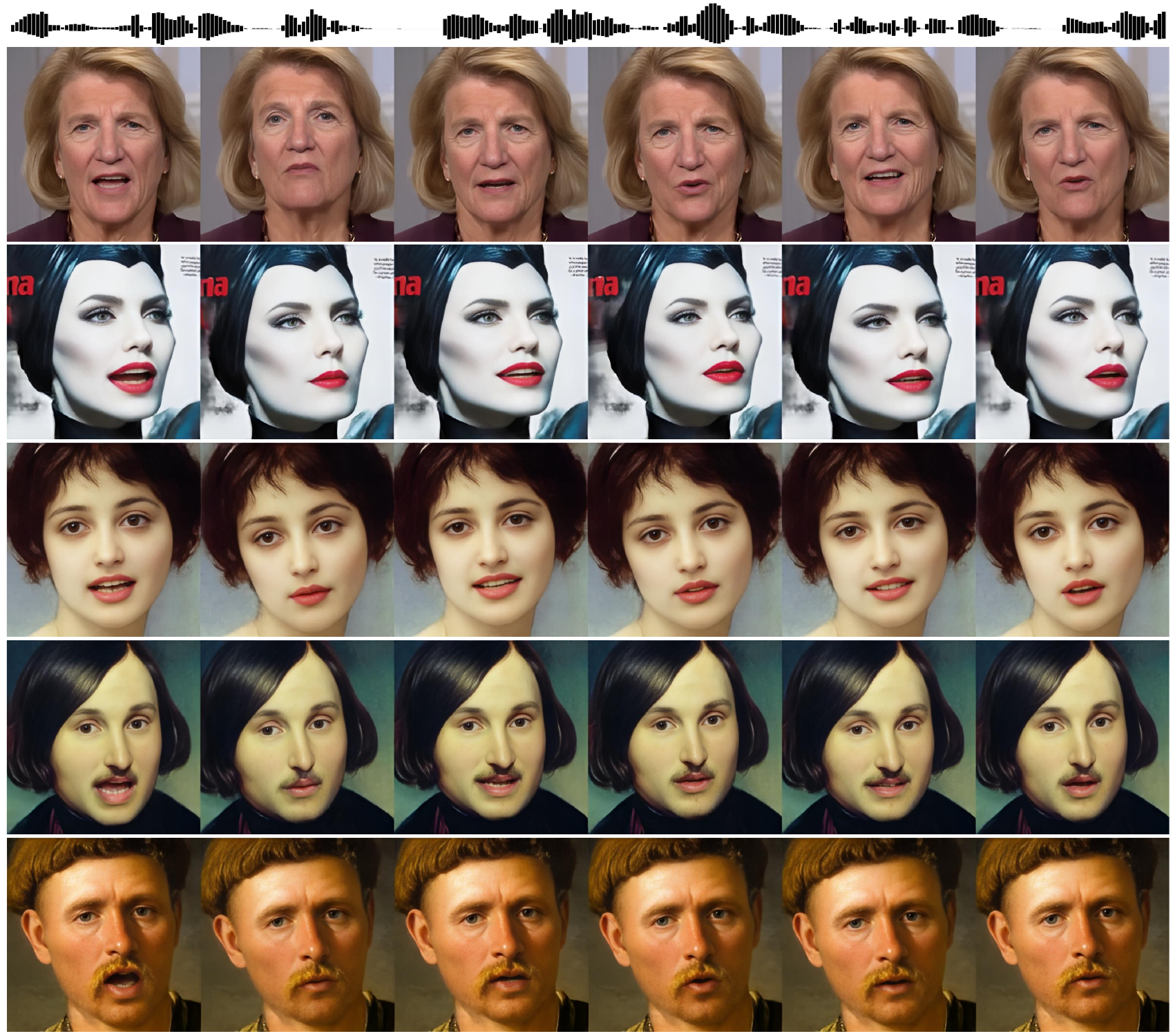}
    \caption{Animation results for out-of-distribution images sharing the same source audio. The first row is the ground-truth video.}
    \label{fig:ood}
\end{figure}

To validate the robustness and generalization capability of our method on out-of-distribution source images, we present some animation results in Fig. \ref{fig:ood}. Upon inspection, our face animator demonstrates consistently good performance across diverse styles and types of input images. These results highlight the effectiveness of our approach in producing natural and expressive talking head animations across varied input conditions, underscoring its potential for real-world applications.

\subsection{One-to-many Generation}
\begin{figure}[t]
    \centering
    \includegraphics[width=\linewidth]{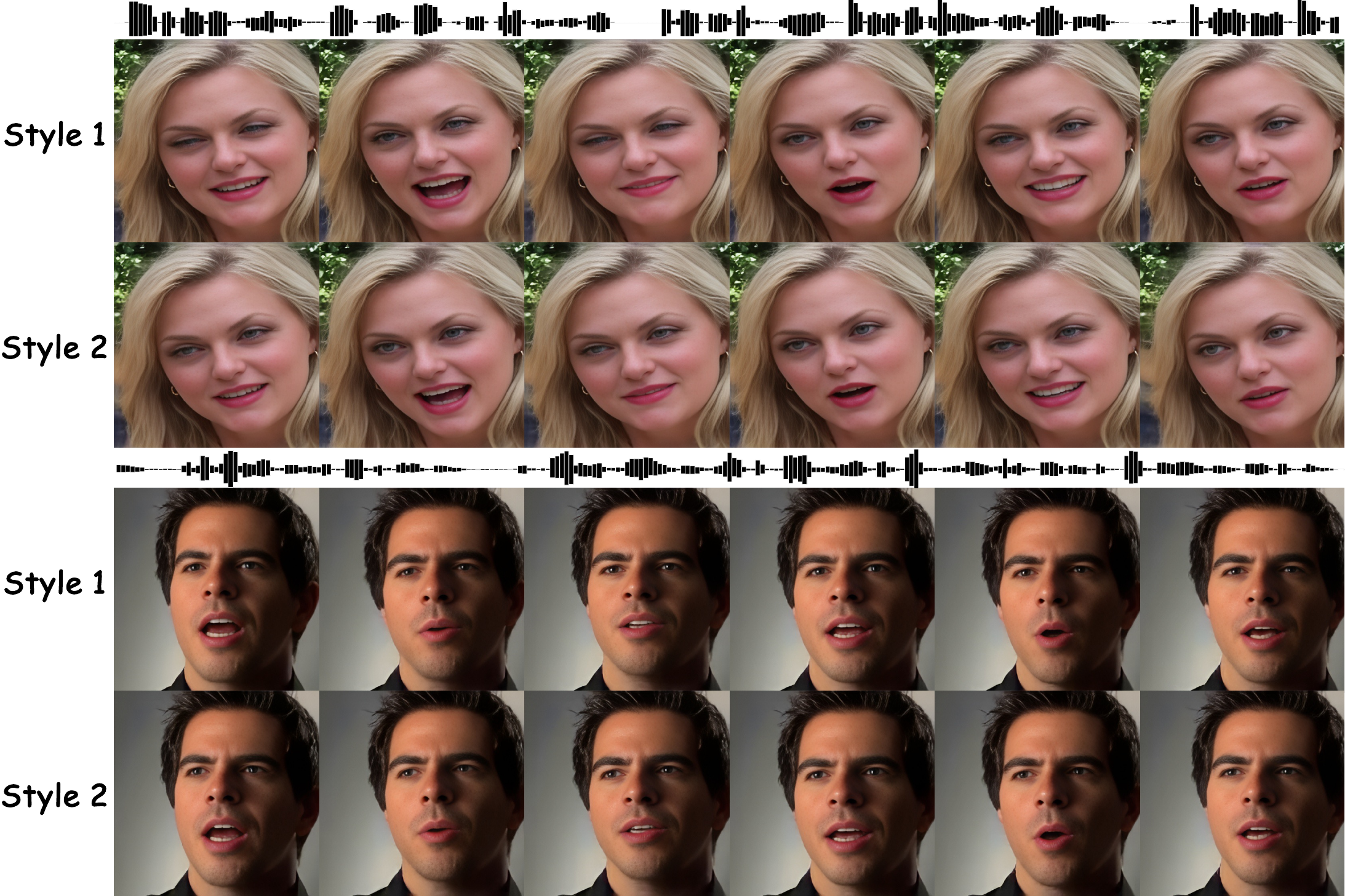}
    \caption{Diverse results for two example identities.}
    \label{fig:diversity}
\end{figure}

We provide additional results in Fig. \ref{fig:diversity} to show the ability of our method in handling one-to-many mappings between audio and non-lip facial motions. In addition to controllable gaze directions, the eye motions and head poses can vary slightly while corresponding to the same audio, verifying the the effectiveness of the proposed method in resolving diverse generation of spontaneous motions.

\section{Supplementary Video}
We created a supplementary video showcasing some of our experimental results and uploaded it to YouTube. For the best viewing experience, we recommend selecting the 720P or 1080P resolution. You can watch the video here: https://youtu.be/c1FsPjHrEI0. 

\section{Limitations and Future Directions}
While our approach achieves notable success in robust and realistic talking face animation, it is crucial to acknowledge certain limitations encountered during our exploration. Even with a latent navigating approach to transform talking motions on source images, our input motion descriptors remain explicit representations that cannot accurately convey subtle head or facial movements. Consequently, in the final driven results, we observe a certain degree of information loss, including a decrease in image resolution and the absence of some details such as wrinkles and changes in teeth. These issues can partly be resolved by off-the-shelf facial restorers such as GFPGAN \cite{wang2021gfpgan} and CTCNet \cite{gao2023ctcnet}. Furthermore, since we adopt a frame-by-frame driving approach to generate the final speaker's facial video, our face animator does not consider temporal continuity, potentially resulting in a slight jitter in the final generated results, somewhat compromising the authenticity of the generated video. These issues will serve as ample motivation for our subsequent research endeavors.

\section{Ethical Considerations} 
Our approach centers on realistic talking face generation that can be applied in digital avatar creation and mixed reality content. Despite potential misuse concerns, we take precautions by embedding visible and invisible video watermarks for content identification, inspired by Dall-E \cite{ramesh2022hierarchical} and Imagen \cite{saharia2022photorealistic}. In addition, we support the forgery detection area by sharing our generated results, assisting in the development of robust algorithms for more intricate scenarios. We underscore the importance of responsible technology use for positive societal impact in machine learning research and daily life.
\end{appendices}

\end{document}